\documentclass{article}

\usepackage{microtype}
\usepackage{graphicx}
\usepackage{subfigure}
\usepackage{booktabs} % for professional tables

\usepackage{multirow}
\usepackage{pifont}
\usepackage[algo2e,ruled,noend]{algorithm2e}

\usepackage{hyperref}

\usepackage[accepted]{icml2024}

\usepackage{amsmath}
\usepackage{amssymb}
\usepackage{mathtools}
\usepackage{amsthm}

\usepackage[capitalize,noabbrev]{cleveref}

\theoremstyle{plain}

\theoremstyle{definition}

\theoremstyle{remark}

\usepackage[textsize=tiny]{todonotes}

\icmltitlerunning{Multimodal Federated Learning with Missing Modality via Prototype Mask and Contrast}

\begin{document}

\twocolumn[
\icmltitle{Multimodal Federated Learning with Missing Modality \\
           via Prototype Mask and Contrast}



\icmlsetsymbol{equal}{*}

\begin{icmlauthorlist}
\icmlauthor{Guangyin Bao}{tj}
\icmlauthor{Qi Zhang}{tj}
\icmlauthor{Duoqian Miao}{tj}
\icmlauthor{Zixuan Gong}{tj}
\icmlauthor{Liang Hu}{tj}
\icmlauthor{Ke Liu}{bnu}
\icmlauthor{Yang Liu}{nudt}
\icmlauthor{Chongyang Shi}{bit}
\end{icmlauthorlist}

\icmlaffiliation{tj}{CEIE, Tongji University, Shanghai, China}
\icmlaffiliation{bnu}{Beijing Normal University, Beijing, China}
\icmlaffiliation{bit}{Beijing Institute of Technology, Beijing, China}
\icmlaffiliation{nudt}{National University of Defense Technology, Hunan, China}

\icmlcorrespondingauthor{Guangyin Bao}{baogy@tongji.edu.cn}
\icmlcorrespondingauthor{Qi Zhang}{zhangqi\_cs@tongji.edu.cn}
\icmlcorrespondingauthor{Duoqian Miao}{dqmiao@tongji.edu.cn}

\icmlkeywords{Federated Learning, Multimodal Learning, Prototype Learning}
\vskip 0.3in
]



\printAffiliationsAndNotice{\icmlEqualContribution} % otherwise use the standard text.

\begin{abstract}

% In federated scenarios, multimodal learning often faces the challenge of uncertain modality missing. Existing methods for addressing missing modalities involve incorporating additional branches into the model or implementing intricate training strategies. However, these methods are not inherent properties of the federated learning framework. In this paper, we introduce a prototype library into the FedAvg-based Federated Learning framework, thereby empowering the framework itself with the capability to alleviate the global model performance degradation resulting from modality missing during both training and testing. The proposed method utilizes prototypes as masks representing missing modalities to formulate a task-calibrated training loss and a model-agnostic uni-modality inference strategy. In addition, a proximal term based on prototypes is constructed to enhance local training. Experimental results demonstrate the state-of-the-art performance of our approach. Compared to the baselines, our method improved training accuracy by 3.7\% with 50\% modality missing and inference accuracy by 23.8\% with uni-modality data.

% In real-world scenarios, multimodal federated learning often faces the challenge of intricate modality missing, which poses constraints on building federated frameworks and significantly degrades model inference accuracy. 
In real-world scenarios, the challenge of random modality missing in multimodal federated learning (mFL) hampers the development of federated frameworks and significantly diminishes model inference accuracy.
% Existing solutions for addressing missing modalities generally involve developing modality-specific encoders on clients and training modality fusion modules on servers. 
% However, these methods are primarily constrained to specific scenarios with either unimodal clients or complete multimodal clients, struggling to generalize effectively in the intricate modality-missing scenarios. 
% Their solutions for addressing missing modalities typically involve creating modality-specific encoders on clients and training modality fusion modules on servers, struggling to generalize effectively in the intricate modality-missing scenarios.
However, existing mFL methods are predominantly limited to specific scenarios with either unimodal clients or complete multimodal clients. They typically create modality-specific encoders on clients and train modality fusion modules on servers, suffering from severe task drift between clients and servers and struggling to generalize effectively in intricate modality-missing scenarios.
In this paper, we introduce a prototype library into the FedAvg-based Federated Learning framework, thereby enabling mFL to alleviate the task drift and performance degradation resulting from modality missing during both training and inference. 
The proposed method utilizes prototypes as masks representing missing modalities to formulate a task-calibrated training loss and a model-agnostic uni-modality inference strategy. 
In addition, a proximal term based on prototypes is constructed to enhance local training. 
Extensive experiments demonstrate the state-of-the-art performance of our approach across a series of missingness settings. 
% Compared to the baselines, our method improved inference accuracy by 3.7\% with 50\% modality missing during training and by 23.8\% during uni-modality inference.
Code is available at \url{https://github.com/BaoGuangYin/PmcmFL}.

\end{abstract}

\section{Introduction}
\label{sec:intro}

Multimodal pre-trained models have exhibited superior performance in various downstream tasks~\cite{wang2023beit3, Yu2022coca, Chen2023pali}, with the availability of large-scale data being a major contributing factor. 
However, collecting large-scale data in practical applications may result in privacy leakage. 
An alternative approach is to adopt a decentralized machine learning paradigm, such as federated learning (FL). 
In FL~\cite{McMahan2017Fedavg}, distributed clients collaborate to train a global model without sharing their private datasets. 
% While FL can provide ample data and efficient communication for training multimodal models, it faces practical challenges when dealing with complex real-world multimodal data. 
However, it faces practical challenges when dealing with complex real-world multimodal data.
One such challenge is the widely-existing issue of missing modalities.
The presence of missing modalities on each client poses constraints on local training thereby leading to a significant drop in global model inference accuracy. 
Furthermore, the non-independent and identically distributed (non-IID) nature of data among clients~\cite{Zhu2021heterFL1,MendietaYW0D2022heterFL2,fedprox,HuangY2022heterFL3,QuZLXW00R2022heterFL4,moon} sharpens the challenges of modality missing, highlighting the urgent need to address the issue in practical multimodal data.
%Specifically, there is a lack of sufficient and accurate prior knowledge of missing modalities 

\begin{figure}[t]
    \begin{center}
    \subfigure[Differences in task scenarios.]{
        \includegraphics[width=1\linewidth]{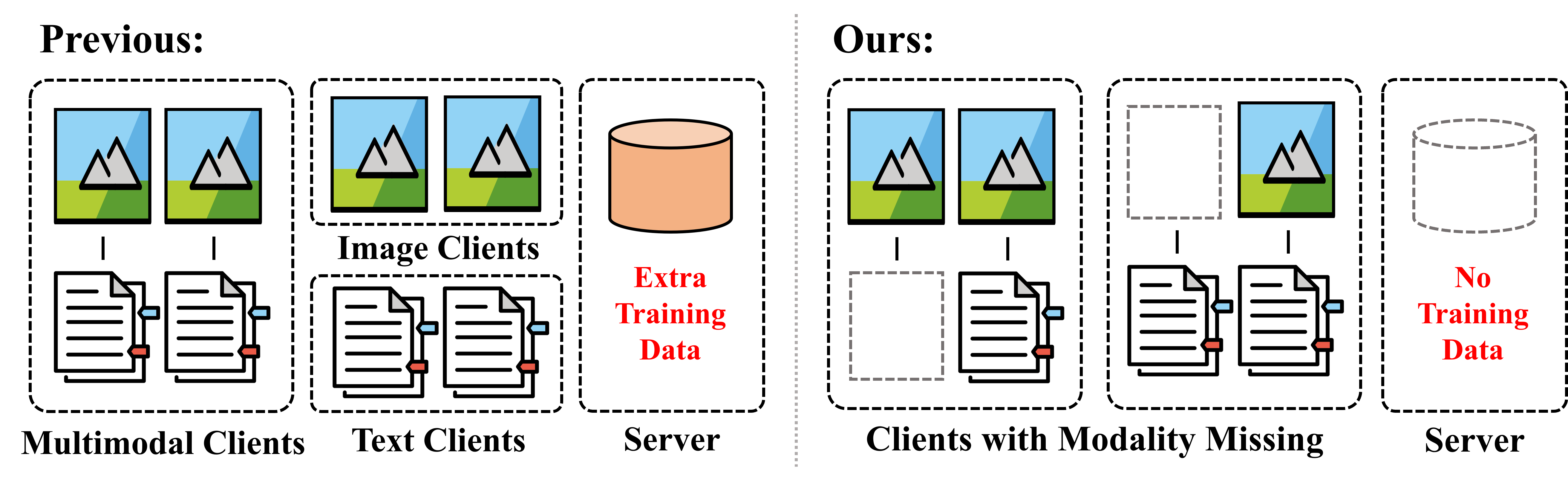}
        \label{difference:scenarios}
        }
    
    \vspace{-4pt}
    \subfigure[Differences in model architectures.]{
        \includegraphics[width=1\linewidth]{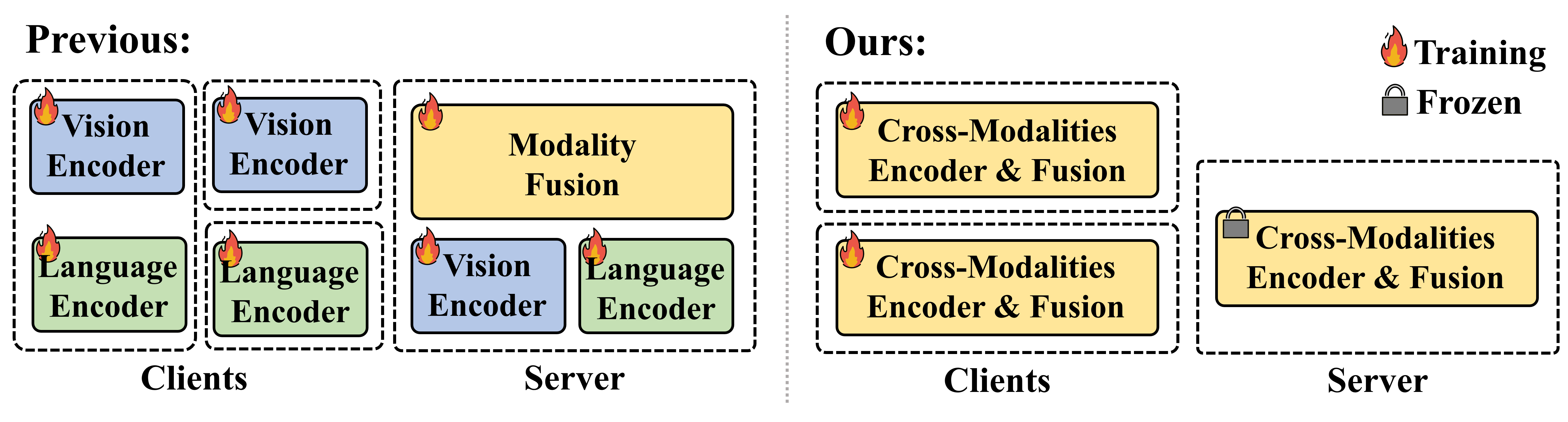}
        \label{difference:architectures}
        }
    \vspace{-8pt}
    \caption{
    Illustration of differences between our work and previous works.
    For \textbf{task scenarios}, previous works consider unimodal clients and modality-complete multimodal clients.
    In contrast, our task scenario involves the possibility of partial modalities being missing on each client.
    In addition, the difference lies in whether the server includes a training set or not.
    For \textbf{model architectures}, previous works train modality-specific encoders for each modality and then train the modality fusion module on the server.
    In contrast, % to avoid extra training on the server, 
    we develop local models with the interactive-encoder architecture encompassing modality encoding and fusion.
    }
    \vspace{-15pt}
    \label{fig:compare}
    \end{center}
\end{figure}

Some pioneer works~\cite{Yu2023creamfl,Zhao2022fediot,chen2022fedmsplit,Feng2023fedmultimodal} attempt to carry out multimodal federated learning with missing modalities. These approaches employ modality-specific encoders for each modality, such as visual encoders for images and language encoders for text, and train them using unimodal or self-supervised training tasks. These local encoders are then either aggregated into a global encoder~\cite{Zhao2022fediot}, or the global encoder is trained by aligning the local encoders through knowledge distillation (KD) on public data~\cite{Yu2023creamfl}.
To perform downstream tasks that require modality fusion, the server needs additional training data to train a modality fusion module and a downstream task head.
These previous works focused on simple scenarios of modality missing, consisting of unimodal clients and modality-complete multimodal clients (as shown in Figure \ref{difference:scenarios}).
However, it is common for each client to have partial data with missing modalities~\cite{Feng2023fedmultimodal}, leading to more intricate practical scenarios. 
An example is observed in social media platforms where users generate three types of data: images only, text only, and image-text pairs.
\vspace{-1pt}

Facing such intricate scenarios of modality missing, the existing multimodal FL frameworks encounter the following issues: 
1) Clients cannot obtain fused representations due to only modality-specific encoders being deployed.
2) Severe task drift occurs between clients and server, e.g., image/text classification on clients but visual question answering (VQA) on the server, resulting in inconsistent optimization directions.
3) Lack of any strategy for addressing modality missing.
4) The availability of the server's downstream task datasets has also been controversial in FL.
These motivate us to propose a new general multimodal FL framework for real-world modality missing and fused representations learning.
\vspace{-1pt}

In this paper, we aim to empower the FL framework with the ability to handle intricate modality-missing scenarios during both training and inference.
Moreover, we develop modality-interactive encoders and fusion modules for each client to avoid the necessity of training data on the server (cf. Figure \ref{difference:architectures}).
To this end, we need to address three challenges: 
CH1. How to handle complex missing patterns for interactive encoder?
CH2. How to perform representation fusion when data is non-IID and modalities are missing?
CH3. How to ensure the performance of the global model when there are missing modalities during inference?
% \vspace{-1pt}

Specifically, we innovatively utilize the flexible Multiway Transformers~\cite{wang2023beit3,beitv2,beit} to construct our multimodal encoders.
By selecting different modality expert networks, Multiway Transformers can serve as both modality-specific encoders and interactive encoders, thereby adapting to complex patterns of modality missing (CH1). 
Inspired by prototype learning in FL~\cite{protoFL1,protoFL2,protoFL3,protoFL4,protoFL5,protoFL6}, we propose a novel prototype-based multimodal FL framework, termed \textbf{PmcmFL} (\textbf{P}rototype \textbf{M}ask and \textbf{C}ontrast for \textbf{M}ultimodal \textbf{FL}) to achieve fused representation learning with non-IID data and missing modalities (CH2). 
In PmcmFL, we construct and maintain a prototype library.
During training, the prototypes act as global prior knowledge of the missing modality to compensate for cross-modal fusion and calibrate task drift. 
The prototype library also introduces a proximal item based on prototypes to reduce heterogeneity among clients.
Consequently, we utilize prototypes for inference with missing modalities (CH3), where various matching algorithms are elaborately introduced to identify the prototype with the closest semantics.

Our experiments show that PmcmFL achieves state-of-the-art performance. 
Notably, it brings 0.2-3.7\% accuracy improvements across different modality missing rates.
Regarding inference with missing modalities, it achieves a remarkable 23.8\% accuracy improvement.

The main contributions are summarized below:

\begin{itemize}
    \item Our work stands out as the first attempt to empower our FL framework with the capability to alleviate the global model performance degradation resulting from modality missing during both training and inference.
    \item We propose PmcmFL to achieve task-calibrated training and higher inference accuracy when dealing with modality-incomplete data.
    \item To the best of our knowledge, we are the first to adopt Multiway Transformer as a versatile encoder to address the complex patterns of modality missing.
\end{itemize}

\section{Related Work}
\label{sec:related_work}

\begin{figure*}[h]
\begin{center}
    \includegraphics[width=0.85\linewidth]{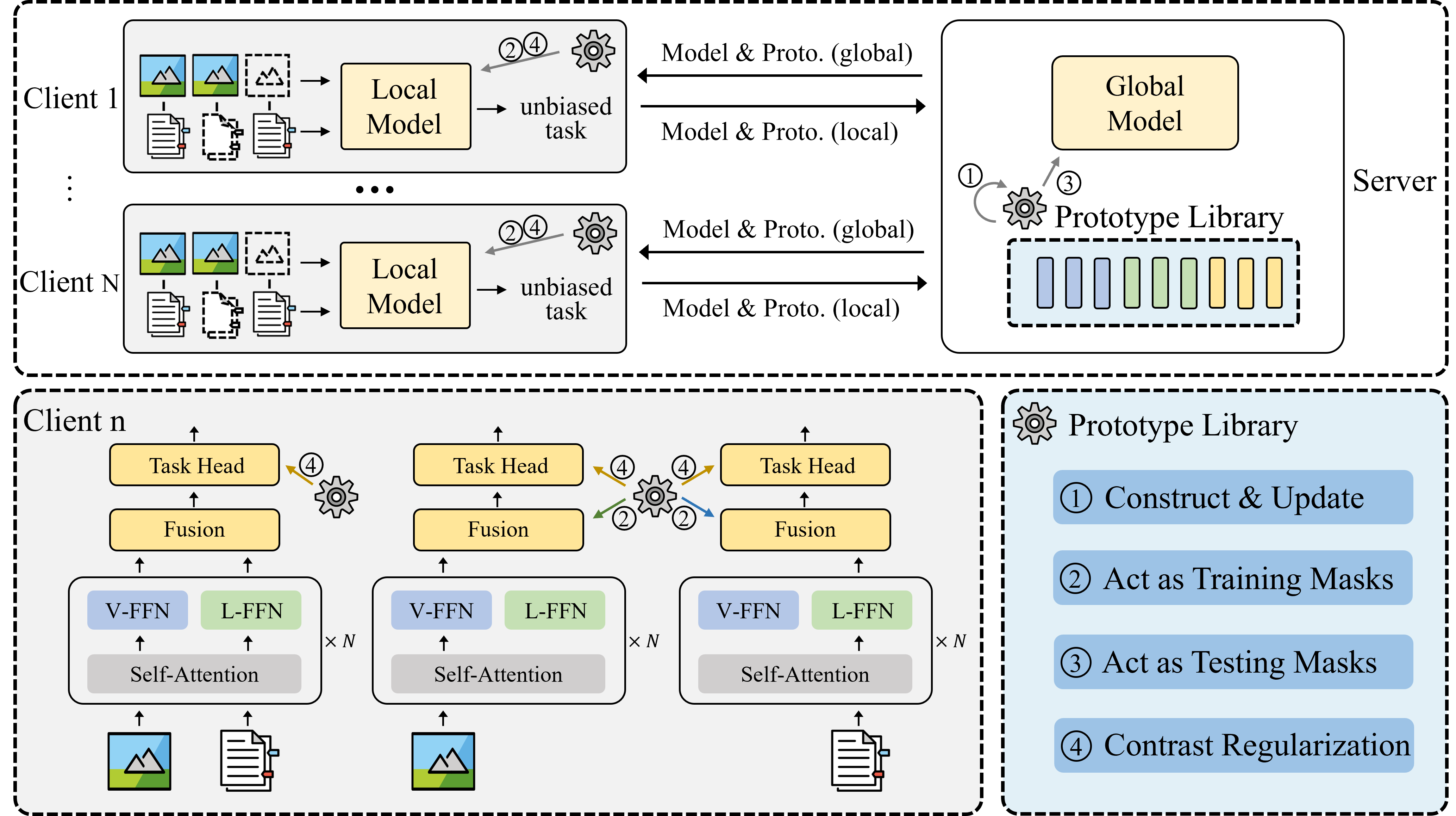}
    \vspace{-1mm}
    \caption{An overview of the proposed PmcmFL. We choose low-dimension representations to construct the prototype library (in Sec.\ref{method:construct}). Based on this, prototypes act as masks to aid training and inference with missing modalities (in Sec.\ref{method:mask}). Besides, we utilize prototypes to construct contrastive loss for better learning of fused representations (in Sec.\ref{method:contrast}).}
    \vspace{-4mm}
    \label{fig2}
\end{center}
\end{figure*}

\subsection{Multimodal Learning}
Multimodal learning has attracted increasing attention from the research community.
The model with dual-encoder architecture~\cite{clip} uses separate encoders for each modality, with shallow modality interaction.
On the contrary, the models with interactive-encoder architecture~\cite{wang2023beit3,vilt} process input from different modalities and concentrate on modeling modality interactions via fuse tokens without modality-specific encoders.

In order to overcome the missing modality issue in multimodal learning, many methods have been developed.
Ma et al.~\cite{MaRZTWP2021SMIL} propose the SMIL to train a feature reconstruction network using a meta-learning algorithm.
Zhao et al.~\cite{MMIN} propose the MMIN to learn robust multimodal representations by training cascade residual autoencoders.
Ma et al.~\cite{MA2022Missing} enhance the robustness of Transformer through multi-task learning and optimal fusion strategy search.
Wang et al.~\cite{WANG2023ShaSpec} propose the ShaSpec to generate features of missing modalities using a shared encoder.
These methods have introduced additional branches and complex training algorithms to the model, which are unsuitable for FL.

\subsection{Federated Learning with Prototype}
Prototype refers to the centroid of the instances belonging to the identical class~\cite{prototype}.
Due to its scalability, prototypes are widely used to solve various problems in FL~\cite{protoFL1,protoFL2,protoFL7,protoFL8}.
Tan et al.~\cite{fedproto} propose FedProto to conduct FL using prototypes rather than aggregation.
Huang et al.~\cite{protoFL3} utilize prototypes to solve domain shift in FL.
Dai et al.~\cite{protoFL4} utilize prototypes to alleviate the performance decline caused by heterogeneous data in FL.
Some other FL frameworks~\cite{protoFL5,protoFL6}, while not explicitly mentioning prototypes, still leverage the concept of prototypes to address corresponding issues.
In this paper, we primarily utilize prototypes to address modality missing issues.

\subsection{Multimodal Federated Learning}
Combining multimodal learning with federated learning is a novel research problem, with the key challenge being how to perform modality interactions when there are missing modalities for each client.
Zhao et al.~\cite{Zhao2022fediot} propose FedIoT to train autoencoders for each modality on every client and give more aggregation weight to multimodal clients.
Chen et al.~\cite{chen2022fedmsplit} propose FedMSplit to construct a dynamic graph for adaptively selecting multimodal client models where some modalities may be missing.
Yu et al.~\cite{Yu2023creamfl} propose CreamFL to conduct cross-modal interaction using inter-modal and intra-modal contrast.
However, these studies only considered scenarios involving either unimodal or modality-complete multimodal clients.
Feng et al.~\cite{Feng2023fedmultimodal} propose FedMultimodal as the first attempt to consider scenarios where modalities of partial data may be missing on each client. 
They address this by masking the missing modalities using zero tensors.
In this paper, we consider scenarios similar to FedMultimodal, with a focus on an efficient and versatile method for addressing modality missing.

\section{Methodology}
\label{sec:methodology}

\subsection{Overview}
\label{method:overview}
We consider a federated learning setting including $N$ multimodal clients.
Without loss of generality, we take images and text as instances of multimodal data.
In Appendix~\ref{appendix: extend}, we analyze how to generalize our framework to other multimodal tasks.
We decouple the model architecture into a representation encoder $\mathcal{E}$, a deep fusion layer $\mathcal{F}$, and a task head $\mathcal{G}$.
In each image-text pair, the image or text will be missing with a probability of $\rho$.
In this case, all the images can be denoted as $\mathcal{I}_{n}=\{(i_{nk},y_{nk})\}_{k=1}^{|\mathcal{I}_{n}|}$, where $i_{nk}$ is the $k$-th image on the $n$-th client, $y_{nk}$ is its corresponding label.
Similarly, all the text can be denoted as $\mathcal{T}_{n}=\{(t_{nk},y_{nk})\}_{k=1}^{|\mathcal{T}_{n}|}$, and all the image-text pairs can be denoted as $\mathcal{M}_{n}=\{(i_{nk},t_{nk},y_{nk})\}_{k=1}^{|\mathcal{M}_{n}|}$.
Furthermore, each client develops a local model $(\mathcal{E}_{n},\mathcal{F}_{n},\mathcal{G}_{n}), n\in [1\dots N]$, with the same architecture as the global model.
\vspace{-1pt}

As shown in Figure \ref{fig2}, PmcmFL constructs or updates the prototype library on the server (operation \ding{172}) at the beginning of each federated communication round.
Subsequently, the prototype library and the global model are broadcast to all participants.
Then, each participating client performs local training and then updates local prototypes.
Using prototypes as masks of missing modalities, a task-calibrated training loss will supervise the training (operation \ding{173}).
Additionally, to tackle the non-IID issue and the resulting client drift, a prototype-based contrastive loss is proposed (operation \ding{175}).
Finally, all participating clients transmit their local models and local prototypes back to the server, after which the global model is updated by aggregating those local models.
During inference, the prototype library also assists inference with missing modalities by transmitting prototypes to the global model (operation \ding{174}).
The complete algorithm is in Appendix~\ref{appendix: algorithm}.
\vspace{-1pt}

For local training, We use Multiway Transformer as encoder $\mathcal{E}$. 
This encoder~\cite{beit,beitv2,wang2023beit3} handles different missing patterns by switching among various encoding modes.
When only image or text input exists, the multiway encoder functions as a modality-specific encoder, like ViT~\cite{vit} or BERT~\cite{bert}.
When inputting an image-text pair, the multiway encoder functions as an interactive encoder, enabling the representation of one modality to fuse information from the other modality.
More details on the Multiway Transformer are in Appendix~\ref{appendix: multiway transformer}.
\vspace{-2pt}

\subsection{Constructing Prototype Library}
\vspace{-1pt}
\label{method:construct}
Prototypes compact the semantics of data within the same class.
Considering communication overhead in FL, low-dimension representations are used to construct the prototype library.
As for the compacting strategy, we naively use the class centroids as the prototypes.
\vspace{-1pt}

\textbf{Selecting Hidden Representations.} \;
For $(i_{nk},t_{nk})\in\mathcal{M}_{n}$, we denote $\mathbb{H}_{nk}=\mathcal{E}_{n}(i_{nk},t_{nk})
\in\mathbb{R}^{m\times d}$ as the output of the multiway transformer encoder, where $m$ is the number of output tokens, and $d$ is the dimension of each token.
The image CLS token $h_{nk}^{(i)}$ and the text CLS token $h_{nk}^{(t)}$ are extracted from $\mathbb{H}_{nk}$.
We select the two CLS tokens along with the fused representation, i.e., $h_{nk}^{(f)}=\mathcal{F}_{n}(h_{nk}^{(i)},h_{nk}^{(t)})$, for constructing prototypes.
\vspace{-1pt}

\textbf{Construction and Update} \;
In the prototype library, we construct and maintain three types of prototypes: image prototypes $\mathcal{P_{I}}$, text prototypes $\mathcal{P_{T}}$, and fusion prototypes $\mathcal{P_{F}}$.
All prototypes are initialized with zero tensors or random tensors. 
Subsequently, after local training in each federated communication round, all participating clients construct local prototypes by computing class centroids. 
Finally, the global prototypes are aggregated from the local prototypes with numbers of client samples as weights.
The construction process is formalized as follows:
\begin{equation}
  \mathcal{P_I}_{n}^{j}=\sum_{k\in C_n^j}\frac{h_{nk}^{(i)}}{|C_n^j|}, 
  \quad 
  \mathcal{P_I}^{j}=\sum_{n\in P_c}\frac{|\mathcal{I}_{n}|}{|\mathcal{D}|}\mathcal{P_I}_{n}^{j},
  \label{equation04}
\end{equation}
\begin{equation}
  \mathcal{P_T}_{n}^{j}=\sum_{k\in C_n^j}\frac{h_{nk}^{(t)}}{|C_n^j|},
  \quad 
  \mathcal{P_T}^{j}=\sum_{n\in P_c}\frac{|\mathcal{T}_{n}|}{|\mathcal{D}|}\mathcal{P_T}_{n}^{j},
  \label{equation05}
\end{equation}
\begin{equation}
  \mathcal{P_F}_{n}^{j}=\sum_{k\in C_n^j}\frac{h_{nk}^{(f)}}{|C_n^j|}, 
  \quad 
  \mathcal{P_F}^{j}=\sum_{n\in P_c}\frac{|\mathcal{M}_{n}|}{|\mathcal{D}|}\mathcal{P_F}_{n}^{j},
  \label{equation06}
\end{equation}
where $C_n^j$ denotes the data of class $j$ on the $n$-th client, $P_c$ denotes the set of participating clients, and $\mathcal{D}$ denotes all data of participating clients.
For $*\in [\mathcal{I,T,F}]$, $\mathcal{P_*}_{n}^{j}$ denotes the local prototypes and ${P_*}^{j}$ denotes the global prototypes.

\subsection{Prototypes as Masks of Missing Modalities}
\label{method:mask}
When missing modalities occur, previous work~\cite{Feng2023fedmultimodal} masks missing data with zero tensors.
In PmcmFL, prototypes are employed as masks for missing modality representations, thereby incorporating global prior knowledge.
Based on this, a task-calibrated training loss and a model-agnostic unimodal inference strategy are proposed.

\textbf{Task-Calibrated Training Loss.} \;
As shown in Figure \ref{fig2}, during local training with complete image-text pairs, the training loss is defined:
\begin{equation}
  L_{task}=\mathcal{L}(\mathcal{G}_n(\mathcal{F}_n(h_{nk}^{(i)},h_{nk}^{(t)})),y_{nk}),
  \label{equation07}
\end{equation}
where $\mathcal{L}$ denotes the task loss function. For classification tasks, cross-entropy is commonly used.
In the case of local training with text missing, a task-calibrated training loss is constructed using the text prototypes as masks:
\begin{equation}
  L_{task}=\mathcal{L}(\mathcal{G}_n(\mathcal{F}_n(h_{nk}^{(i)},\mathcal{P_T}^j)),y_{nk}),
  \label{equation08}
\end{equation}
where $\mathcal{P_T}^j$ has the same class as $y_{nk}$. 
Similarly, the task-calibrated training loss is defined when images are missing:
\begin{equation}
  L_{task}=\mathcal{L}(\mathcal{G}_n(\mathcal{F}_n(\mathcal{P_I}^j,h_{nk}^{(t)})),y_{nk}).
  \label{equation09}
\end{equation}

With supervision from the task-calibrated loss, the entire model is trained on the original multimodal task, even when there are missing modalities on clients. On the contrary, using zero masks will result in a task-drifted loss. 
For example, in the case where text is missing, the training loss with zero masks is denoted as $\mathcal{L}(\mathcal{G}_n(\mathcal{F}_n(h_{nk}^{(i)},\mathbf{0})),y_{nk})$. 
Essentially, this represents an unimodal training task that depicts the relationship between the unimodal representation $h_{nk}^{(i)}$ and the ground truth $y_{nk}$,
which leads to inconsistent optimization directions for the model training.

\textbf{Model-Agnostic Unimodal Inference.} \;
During inference with missing modalities, prototypes are employed as masks to assist the global model.
Unlike training, where labels are available, we must find corresponding cross-modal prototypes according to embedded representations.
To this end, a matching function $m(\cdot)$ is established from hidden representations to prototypes.
We use an example to describe the matching process of cross-modal prototypes.
Without loss of generality, assume that the model only utilizes images for inference.
First, the image $i_{nk}$ is input for computing the image representation $h_{nk}^{(i)}$.
Sequentially, the image prototype with the same class $\mathcal{P_I}^j$ is determined using this matching function $\mathcal{P_I}^j=m(h_{nk}^{(i)})$.
Following this, the corresponding text prototype $\mathcal{P_T}^j$ is determined according to class $j$.
Finally, the image representation $h_{nk}^{(i)}$ and the text prototype $\mathcal{P_T}^j$ are input into the subsequent modules.

As for matching function $m(\cdot)$, we propose both model-free and model-based matching.
Model-free matching involves matching the closest prototype for each representation using a distance metric, such as L1, L2 distance, or cosine similarity.
And model-based matching involves training tiny models, treating the search of cross-modal prototypes as either a classification or a retrieval task.
These tiny models are trained only in the final round of federated communication, incurring negligible overhead to the FL framework.
For the classification task, we train a shallow MLP-Classifier using cross-entropy loss.
For the retrieval task, inspired by DALLE$\cdot$2~\cite{dalle2}, we sequentially use CLIP loss~\cite{clip} and MSE loss to train an MLP-Prior to achieve transfer and retrieval.
We attempt three aggregation strategies for tiny models from different clients: selecting the model with maximum client samples, model aggregation, and model ensemble.

Inspired by MixUp~\cite{mixup}, we propose ProtoMix to enrich the semantic information within prototypes used for inference. ProtoMix linearly combines the top-$k$ related prototypes with weights derived from applying the Softmax function to distance metrics, classification probabilities, or retrieval match scores. 

% Our inference strategy is model-agnostic, as it is a function of our FL framework without modifying the network.% architecture.

\subsection{Prototypes as Learned Representation Targets}
\label{method:contrast}
Due to the non-IID issue, the learned representation distributions in latent space significantly differ among clients (i.e., client drift), leading to an uncoordinated model aggregation.
To achieve model aggregation without conflicts, all clients must theoretically have similar local representation distributions.
In PmcmFL, prototypes are used as learned representation targets to guide clients in learning fused representation distributions clustered by class.
For this reason, we introduce a proximal term based on the unidirectional CLIP contrastive loss to regularize client training.

Sampling a batch $\mathcal{B}$ from the local dataset, the proximal term can be denoted as:
\begin{equation}
  L_{contr}=-\frac{1}{|\mathcal{B}|}
  \sum_{s=1}^{|\mathcal{B}|} \log \frac
  {\exp(h_s^{\top}\cdot\mathcal{P}_s/{\tau})}
  {\sum_{i=1}^{|\mathcal{B}|} \exp({h_s^{\top}\cdot\mathcal{P}_i}/{\tau})},
  \label{equation10}
\end{equation}
where $h_s$ is the $s$-th representation in batch, $\mathcal{P}_s$ is its corresponding prototype, and $\tau$ is a temperature factor.
As the global model gradually converges, the prototypes converge step by step.
This proximal term works by pulling representations closer to the prototypes of the same class and pushing representations away from prototypes of different classes so that the learned representations are around the corresponding global prototypes.

Since prototypes are the centroids of class representations, maintaining a certain distance (i.e., fine-grained semantic difference) from each representation, we use CLIP loss to constrain their similarities rather than relying on MSE to constrain their distances.
Although the prototype-based proximal term can theoretically be constructed on image representations, text representations, and fused representations separately, we only utilize the last one due to the difficulty in coordinating multiple proximal terms.
In this case, the total training loss for each client is the sum of task loss and contrastive loss of fused representations:
\begin{equation}
  L=L_{task}+\gamma L_{contr}^f,
  \label{equation11}
\end{equation}
where $\gamma$ is the weighting hyper parameter for $L_{contr}^f$.

\section{Experiment}
\label{sec:experiment}

\begin{table*}[h]
  \caption{Comparison of PmcmFL with baselines under various missing rates.
  Modality missing occurs only in the training phase, and the data in the inference phase is modality-complete.
  * denotes the PmcmFL's performance is significantly better than existing baselines (paired t-test, $p<0.05$). 
  RM denotes Random Mask.}
  \vspace{-1mm}
\begin{center}
\scalebox{0.9}{
    \begin{tabular}{@{}lcccccc@{}}
    \toprule
    \multirow{2}{*}{Methods} & \multicolumn{5}{c}{Accuracy (\%) on 5K Testing Samples ($\rho_{test}=0$)} & \multirow{2}{*}{\quad Acc@sum \quad} \\
    \cline{2-6}
    & {$\rho_{train}=10\%$} & {$\rho_{train}=20\%$} & {$\rho_{train}=30\%$} & {$\rho_{train}=40\%$g} & {$\rho_{train}=50\%$} & \\
    \midrule
    FedAvg + ignore     & 56.444 & 53.936 & 49.990 & 45.024 & 38.898 & 244.292 \\
    FedMultimodal          & 55.360 & 52.804 & 49.246 & 46.378 & 41.034 & 244.822 \\
    FedMultimodal + RM        & 55.624 & 52.542 & 48.164 & 47.288 & 41.346 & 244.964 \\
    FedProx + Mask                 & 52.506 & 49.546 & 46.502 & 46.862 & 41.140 & 236.556 \\
    FedIoT + Mask                     & 55.010 & 52.180 & 47.558 & 47.582 & 41.838 & 244.168 \\
    FedPAC + Mask                      & 55.028 & 50.742 & 46.916 & 45.930 & 40.242 & 238.858 \\
    FedHKD + Mask                      & 56.286 & 53.546 & 50.544 & 47.832 & 43.376 & 251.584 \\
    \midrule
    PmmFL(\textbf{Ours}) + FA            & 56.144 & 54.366 & 50.556 & 48.796 & 46.168 & 256.030 \\
    PmmFL(\textbf{Ours}) + HKD           & 56.458 & 53.926 & 51.568 & 49.472 & 46.394 & 257.818 \\
    PmcmFL (\textbf{Ours})               & \textbf{ 56.636*} & \textbf{ 55.478*} & \textbf{ 52.256*} & \textbf{ 49.548*} & \textbf{ 47.042*} & \textbf{ 260.960*} \\
    \bottomrule
  \end{tabular}
  }
  \label{main results}
  \vspace{-5mm}
\end{center}
\end{table*}

\subsection{Experiment Setup}
\textbf{Dataset.} \;
Following CreamFL~\cite{Yu2023creamfl}, we utilize the challenging VQAv2 dataset~\cite{vqav2} to evaluate our PmcmFL.
For the efficiency of the experiments, we construct a tiny VQA task, one-tenth the scale of the original VQA task.
The tiny VQA is a classification task of 310 classes, with 64,000 training samples and 5,000 testing samples.
The training data is distributed to 30 clients, employing the Dirichlet distribution ($\alpha=0.1$) for non-IID data partition~\cite{noniid}. 
Following suggestions from previous work~\cite{Feng2023fedmultimodal}, we set an equal training missing rate of $\rho_{train}=[0.1, 0.2, 0.3, 0.4, 0.5]$ for each modality.
More details on the dataset are in Appendix~\ref{appendix: dataset}.

\textbf{Model.} \;
For the encoder $\mathcal{E}$, we utilize the same architecture as BEIT3-base~\cite{wang2023beit3} along with its pre-train parameters.
For the deep fusion layer $\mathcal{F}$, the image CLS token and the text CLS token are concatenated and projected through an MLP.
For the task head $\mathcal{G}$, the fusion representation is projected to logits using another MLP.
More details on the model architecture are in Appendix ~\ref{appendix: model}.
\vspace{-1pt}

\textbf{Baseline.} \;
We compare our PmcmFL with the most relevant existing FL frameworks. 
Among them, 1) FedAvg+ignore, where training is conducted only on modality-complete data, ignoring data with modality missing; 2) FedMultimodal~\cite{Feng2023fedmultimodal}, Essentially an extension of FedAvg, using Zero Mask for missing modalities; 3) Fedmultimodal+random mask, where replaces Zero Mask with Gaussian Random Mask; 4) FedIoT+mask, a framework the same as FedAvg, where Mask is applied to missing modalities, using aggregation strategy is designed for multimodal FL in FedIoT~\cite{Zhao2022fediot}. 5) FedPAC+mask, a framework based on FedAvg and Mask for missing modalities, using 
prototype-based feature alignment (FA) in FedPAC~\cite{protoFL5}. 6) FedHKD+mask, a framework based on FedAvg and Mask for missing modalities, using prototype-based hyper knowledge distillation (HKD) in FedHKD~\cite{protoFL6}.
To demonstrate the effectiveness of our Prototype Contrast, we ablate it to obtain the variant PmmFL(PmcmFL without Contrast), and introduce the following two baselines:
7) PmmFL+FA and 8) PmmFL+HKD, which replace Prototype Contrast with Feature Alignment (FA) and Hyper Knowledge Distillation (HKD), respectively.
We future compare with 9) CreamFL in simple modality missing scenarios.
More details on baselines are in Appendix ~\ref{appendix: baselines}.
\vspace{-1pt}

\textbf{Implement.} \;
To enable fair comparisons, our experiments remain consistent on method-agnostic hyperparameters, including federated communication rounds, local training epochs, client selection rate, learning rate, optimizer parameters, and batch size.
For hyperparameters introduced by PmcmFL, to prevent them from overfitting to specific tasks, we only conducted a simple parameter search.
Only the weight $\gamma$ is searched within $[5.0, 1.0, 0.5, 0.1, 0.01]$ to find the optimal value. 
% For inference with missing modalities, we set the missing rate of one modality to 100\% for intuitive comparison.
% For top-$k$ ProtoMix, we attempt all possible values for $k$.
More Implementation details for all experiments can be found in Appendix~\ref{appendix: implementation}.
\vspace{-1pt}

\textbf{Evaluation Metric.} \;
We fix the random seed and utilize optimal accuracy in the last ten communication rounds as the final performance.
Note that all evaluations are conducted with modality-complete testing samples, except in experiments on inference with missing modalities (Sec. \ref{sec: unimodal inference}).
Appendix~\ref{appendix:Inference with Partial Missing} provides accuracy under varying degrees of modality missing during the inference phase.

\begin{table*}[!t]
  \caption{Results of ablation studies under various missing rates. PC denotes Prototype Contrast and PM denotes Prototype Mask.}
  \begin{center}
  \begin{tabular}{@{}cccccccc@{}}
    \toprule
    \multirow{2}{*}{PC} & \multirow{2}{*}{PM } & \multicolumn{5}{c}{Accuracy (\%) on 5K Testing Samples ($\rho_{test}=0$)} & \multirow{2}{*}{\quad Acc@sum \quad} \\
    \cline{3-7}
    & & {$\rho_{train}=10\%$} & {$\rho_{train}=20\%$} & {$\rho_{train}=30\%$} & {$\rho_{train}=40\%$} & {$\rho_{train}=50\%$} & \\
    \midrule
               &            & 55.624 & 52.542 & 48.164 & 47.288 & 41.346 & 244.964 \\
    \checkmark &            & 55.948 & 53.588 & 50.982 & 47.422 & 44.454 & 252.394 \\
               & \checkmark & 55.326 & 54.486 & 50.858 & 49.290 & 45.848 & 255.808 \\
    \checkmark & \checkmark & \textbf{56.636} & \textbf{55.478} & \textbf{52.256} & \textbf{49.548} & \textbf{47.042} & \textbf{260.960} \\
    \bottomrule
  \end{tabular}
  \vspace{-7mm}
  \end{center}
  \label{ablation}
\end{table*}

\begin{table}[!t]
  \caption{
  Inference with 100\% text modality missing. 
  Zero Mask and Random Mask are baseline methods for handling inference with missing modalities. 
  Others are our Prototype Mask with various matching algorithms.
  % Our Prototype Mask under Top1 ProtoMix, based on classifier matching with aggregation, achieves the best inference accuracy (27.502\%).
  }
  \vspace{-3mm}
\begin{center}  
  \begin{tabular}{@{}lcccc@{}}
    \toprule
    \multirow{2}{*}{Methods} & \multicolumn{4}{c}{Accuracy (\%) on Unimodal Data}  \\
    \cline{2-5}
    & {Mix-1} & {Mix-10} & {Mix-20} & {Mix-best}\\
    \midrule
    Zero Mask               & 0.996  & 0.996 & 0.996 & 0.996 \\
    Random Mask             & 3.662  & 3.662 & 3.662 & 3.662 \\
    \midrule
    L1 Metric               &  6.358 &  6.170 &  6.784 &  6.784 \\
    L2 Metric               &  5.454 &  5.238 &  5.782 &  6.106 \\
    COS Metric              &  5.524 &  6.062 &  6.676 &  6.938 \\
    Classifier(max.)        & 26.956 & 27.022 & 27.022 & 27.022 \\
    Classifier(ens.)        & 27.090 & 27.090 & 27.090 & 27.090 \\
    Classifier(avg.)        & \textbf{27.502} & 27.092 & 27.092 & \textbf{27.502} \\
    MLP-Prior(max.)          & 27.052 & 19.906 &  8.742 & 27.052 \\
    MLP-Prior(ens.)          & 27.260 & 27.280 & 27.280 & 27.280 \\
    MLP-Prior(avg.)          &  5.858 &  5.702 &  5.996 &  6.210 \\
    \bottomrule
  \end{tabular}
   \vspace{-6mm}
\end{center}
\label{unimodal inference}
\end{table}

\subsection{Main Results}
Table \ref{main results} displays the accuracy of the global model in all baselines and our PmcmFL.
Our PmcmFL framework generally achieves noticeable performance improvement over all baselines in all modality missing rates.

Compared to ignoring incomplete data, the Mask method significantly improves the performance of federated training with 50\% modality missing (a 2.448\% boost), indicating that Mask is effective in handling severe modality missing.

Compared to FedAvg, FedIoT leads to a performance decline (a 0.796\% drop on Acc@sum), which indicates that assigning more aggregation weight to multimodal clients no longer works in our task scenarios.
Similarly, FedPAC leads to a performance decline (a 6.106\% drop on Acc@sum), which is because the feature alignment (FA) employed by FedPAC, a strong regularization that narrows the distance between representations and prototypes through MSE, disrupts the semantic distinctiveness of representations.

Compared to all the baselines, our PmcmFL achieves superior performance (9.376-24.404\% boost on Acc@sum), demonstrating the effectiveness of our Prototype Mask and Contrast.
Furthermore, as the modality missing rate increases, the performance improvement brought about by our PmcmFL tends to be more significant (a 0.192\% boost in 10\% missing but a 3.666\% boost in 50\% missing). 

Compared to PmcmFL's variants, slightly superior performance (3.14-4.93\% boost on Acc@sum) indicates that our Prototype Contrast is more generalizable than Feature Alignment and Hyper Knowledge Distillation.
We further validate the better performance of PmcmFL compared with CreamFL in scenarios with a 100\% missing rate of either modality.
The experimental results are in Appendix~\ref{appendix: creamfl}.

\begin{figure}[!t]
    \begin{center}
    \vspace{3mm}
    \subfigure[Representation distribution learned by different clients. Dots of the same color denote representation distribution for one client.]{
        \includegraphics[width=\linewidth]{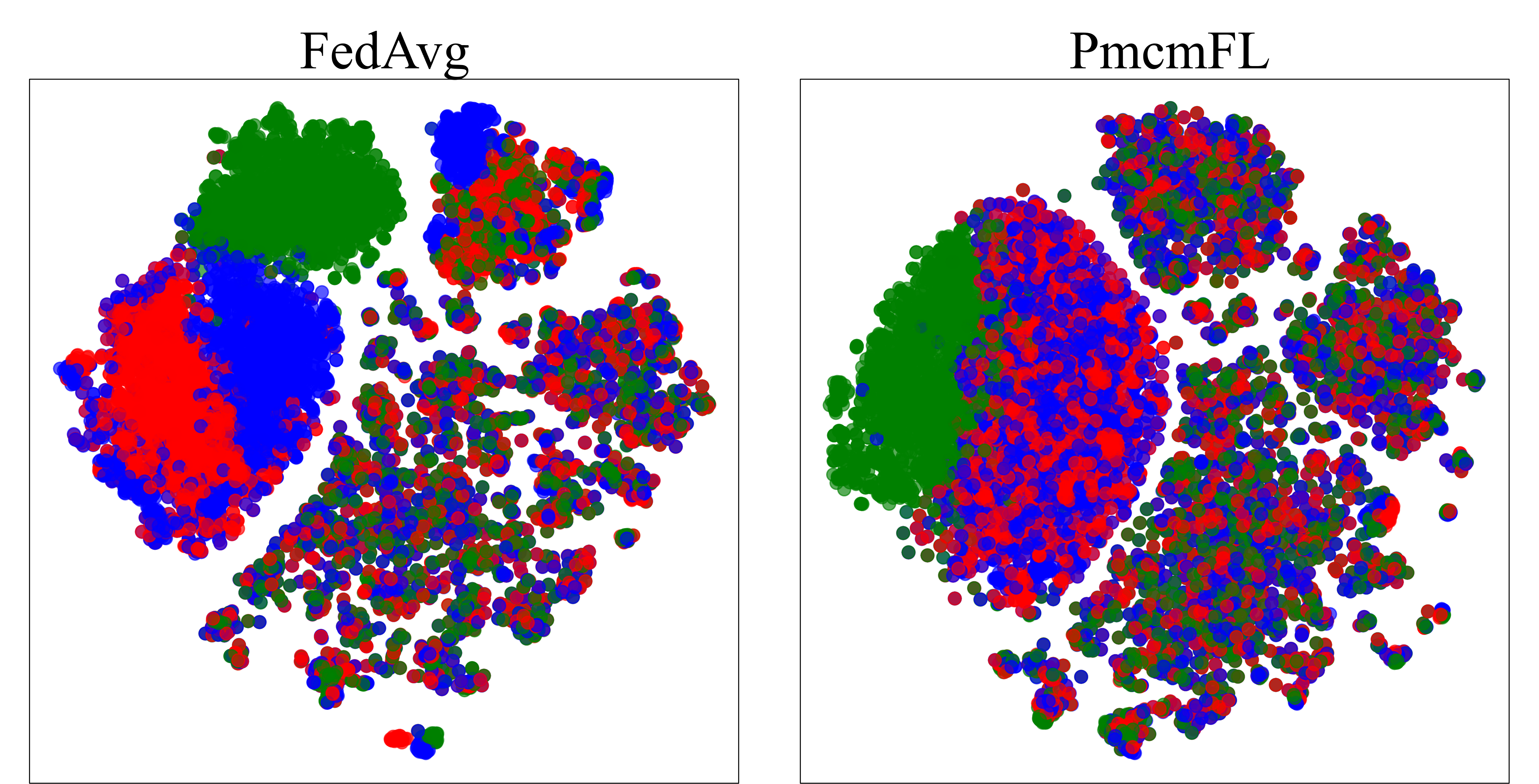}
        \label{tsne for clients}
    }
    \vspace{-2mm}
    
    \subfigure[Representation distribution learned by the global model. Dots of the same color denote representations of the same class. The stars represent prototypes for corresponding classes.]{
        \includegraphics[width=\linewidth]{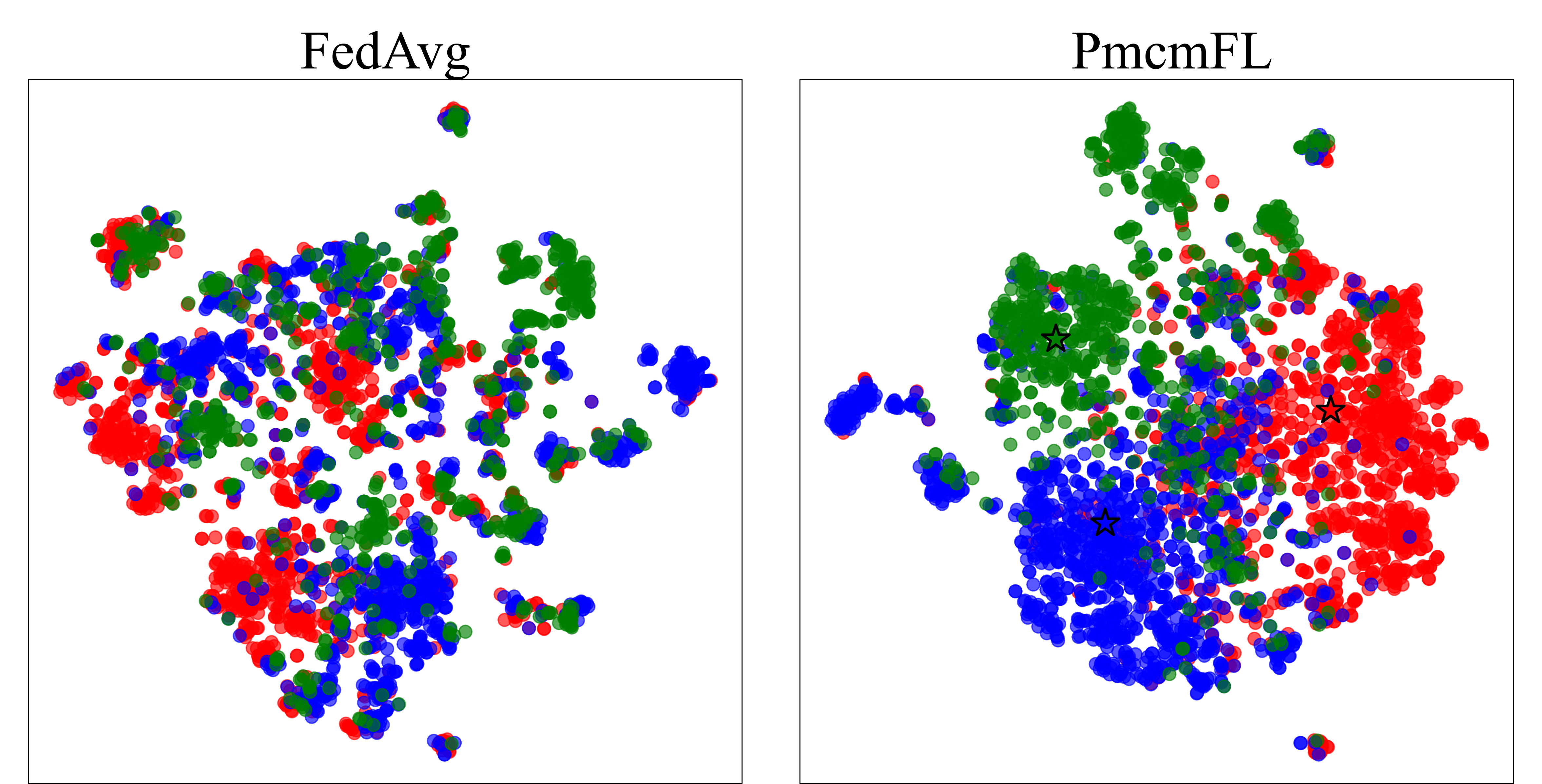}
        \label{tsne for labels}
    }
    \vspace{-5mm}
    \end{center}
    \caption{T-SNE visualization for qualitative studies.}
    \vspace{-6mm}
\label{tsne}
\end{figure}

\subsection{Ablation Studies}
We study the effect of each PmcmFL component during training.
Table \ref{ablation} displays the results of ablation studies under various training modality missing rates.

The results show that Prototype Contrast and Mask contribute to performance improvement over the baselines, with gains of 7.430\% and 11.168\% on Acc@sum, respectively.
Moreover, combining Prototype Contrast and Prototype Mask assist local training yields further performance boost.

\subsection{Inference with Missing Modalities}
\label{sec: unimodal inference}
Table \ref{unimodal inference} shows the testing accuracy of the global model in 100\% text missing, demonstrating intuitively the performance of PmcmFL in handling inference with missing modalities.
We use the global model trained with 30\% modality missing, and its accuracy is 52.256\% in modality-complete testing samples.
For ProtoMix, top-1, top-10, top-20, and best prototype mix are exhibited.

We can obviously see that the inference accuracy of Zero Mask and Random Mask, serving as baselines, is only 0.996\% and 3.662\%, respectively, which indicates that the modality missing issues severely impair the model's inference accuracy.
In contrast, Prototype Mask achieves the highest accuracy of 27.502\% (a 23.840\% boost), demonstrating that our method helps restore modal inference accuracy.
Moreover, the experiments also demonstrate the effectiveness of our ProtoMix strategy.
More results and discussions can be found in Appendix ~\ref{appendix: protomix}.

\subsection{Qualitative Studies}
We utilize T-SNE~\cite{tsne} for visualization to qualitatively analyze the role of Prototypes Contrast in our FL framework.

Figure \ref{tsne for clients} visualizes fused representation distribution learned by different clients.
% All testing samples are used in this figure.
We can observe that in FedAvg, some dots cluster by clients, indicating that each client forms its individual latent space.
When local models are aggregated into a global model, individual latent spaces hinder the formation of a unified latent space.
In contrast, in our PmcmFL, the heterogeneity of latent spaces among various clients is reduced due to prototypes acting as representation targets to guide local training.

Figure \ref{tsne for labels} visualizes the fused representation distribution learned by the global model.
We randomly selected samples of three classes in the testing set.
It can be observed that in FedAvg, dots from different classes are mixed, which is not conducive to subsequent classification tasks. 
In contrast, in our PmcmFL, dots from different classes cluster around their respective prototypes, which validates our starting point of using prototypes for constructing contrastive loss.

\begin{figure}[t]
    \begin{center}
    \subfigure[Communication Overhead]{
        \includegraphics[width=0.43\linewidth]{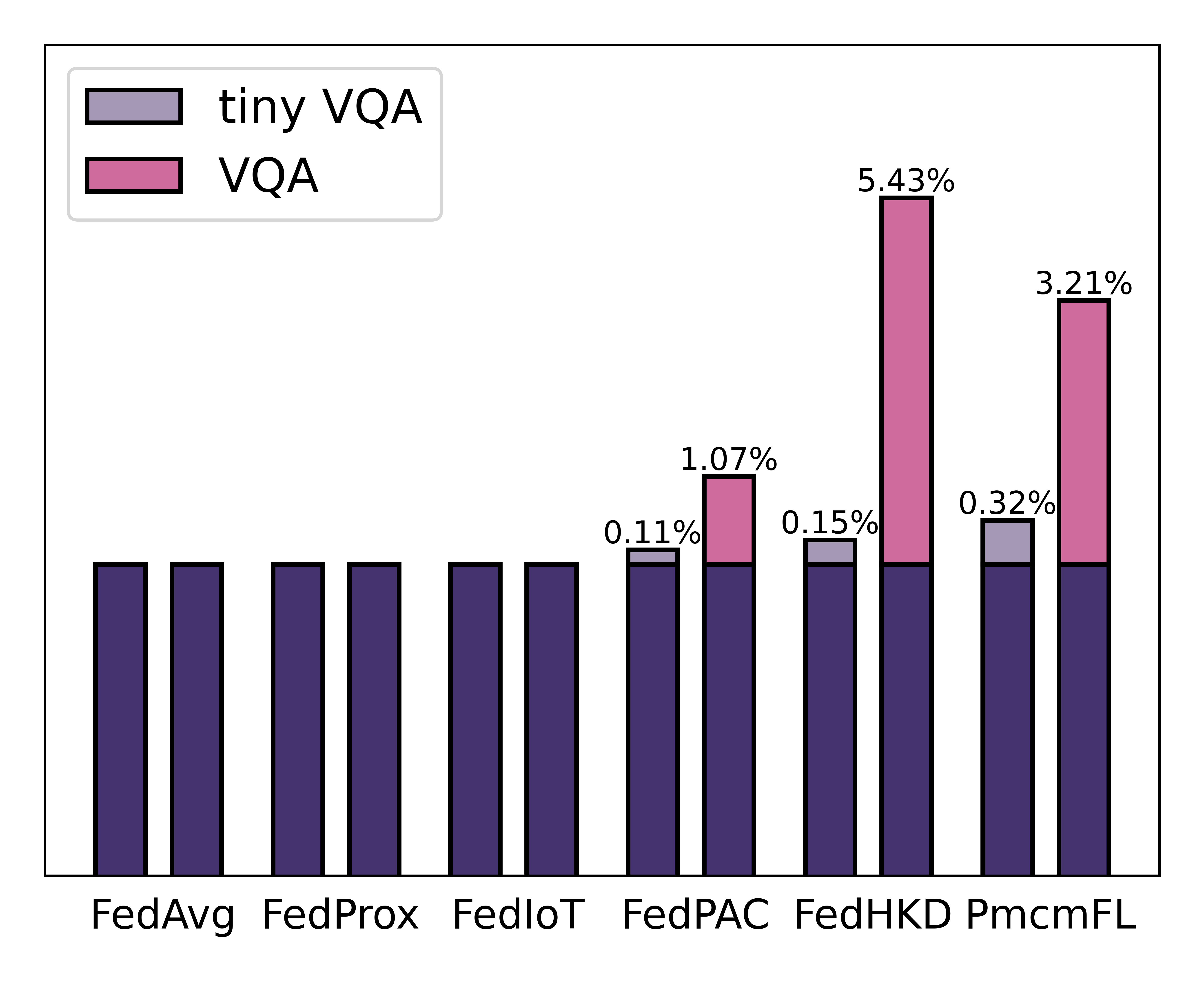}
        \label{overhead}
    }
    \subfigure[Communication Efficiency]{
        \includegraphics[width=0.47\linewidth]{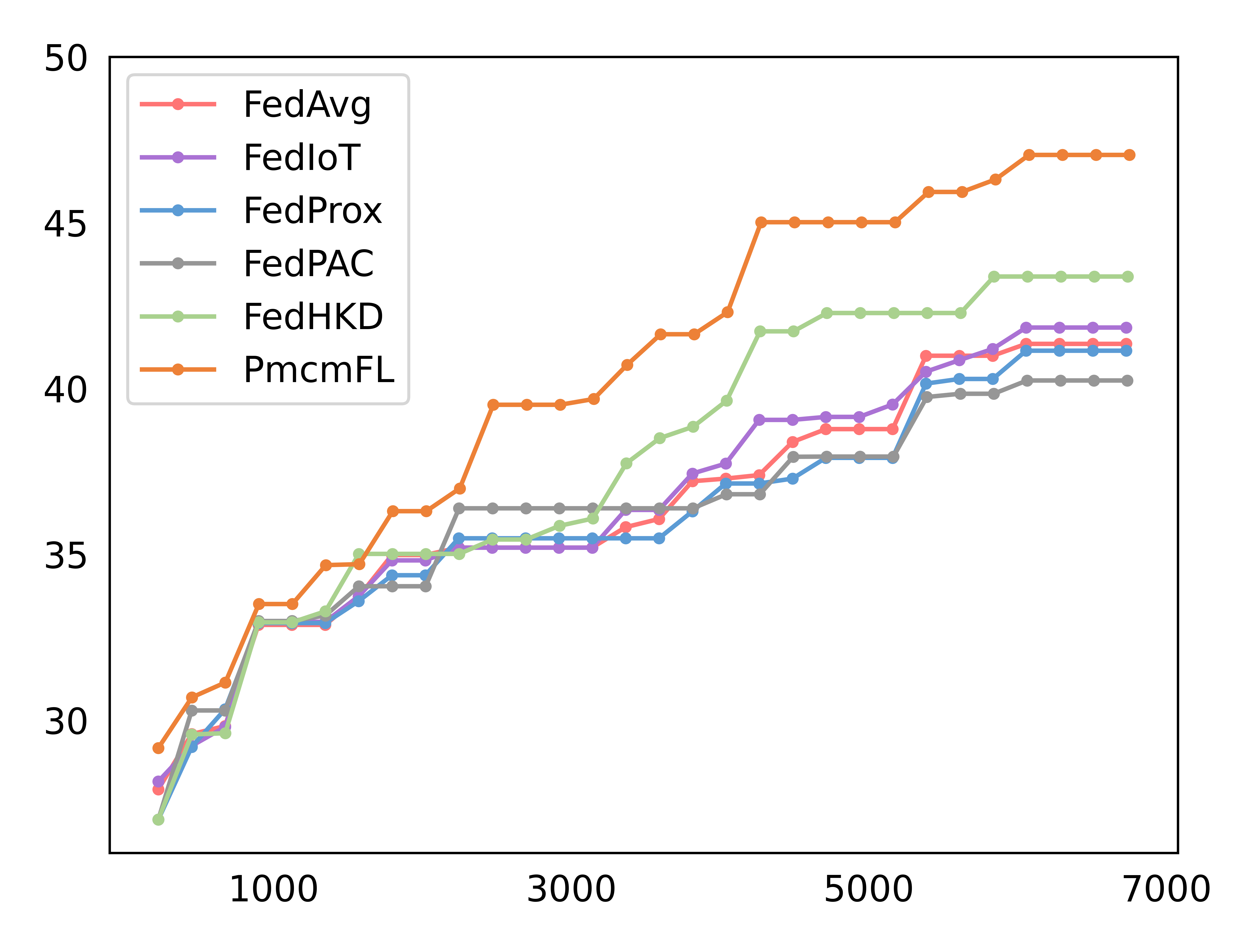}
        \label{efficiency}
    }
    \end{center}
    \vspace{-3mm}
    \caption{Communication overhead and communication efficiency in six FL frameworks. Figure (a) shows the amount of parameters transmitted by different frameworks in one communication round. Figure (b) shows the optimal performance (\%) of the global model when transmitting different parameters (MB).}
    \vspace{-5mm}
\label{commu}
\end{figure}

\subsection{Overhead and Efficiency of Communication}

In our PmcmFL, the parameters transmitted during each federated communication round include model weights and the prototype library.

Figure \ref{commu} illustrates different FL frameworks' communication overhead and communication efficiency.
As shown, in our tiny VQA task, PmcmFL introduces only negligible additional communication overhead (+0.32\%).
In the original VQA task, PmcmFL exhibits even lower communication overhead than FedHKD (+3.21\% v.s. +5.43\%).
Furthermore, we can observe that when transmitting the same amount of parameters, PmcmFL achieves optimal performance, demonstrating that our framework has the highest communication efficiency.

\subsection{Robustness to federated settings}
\label{Robustness to federated settings}

\begin{figure}[t]
    \begin{center}
    \subfigure[Communication rounds]{
        \includegraphics[width=0.46\linewidth]{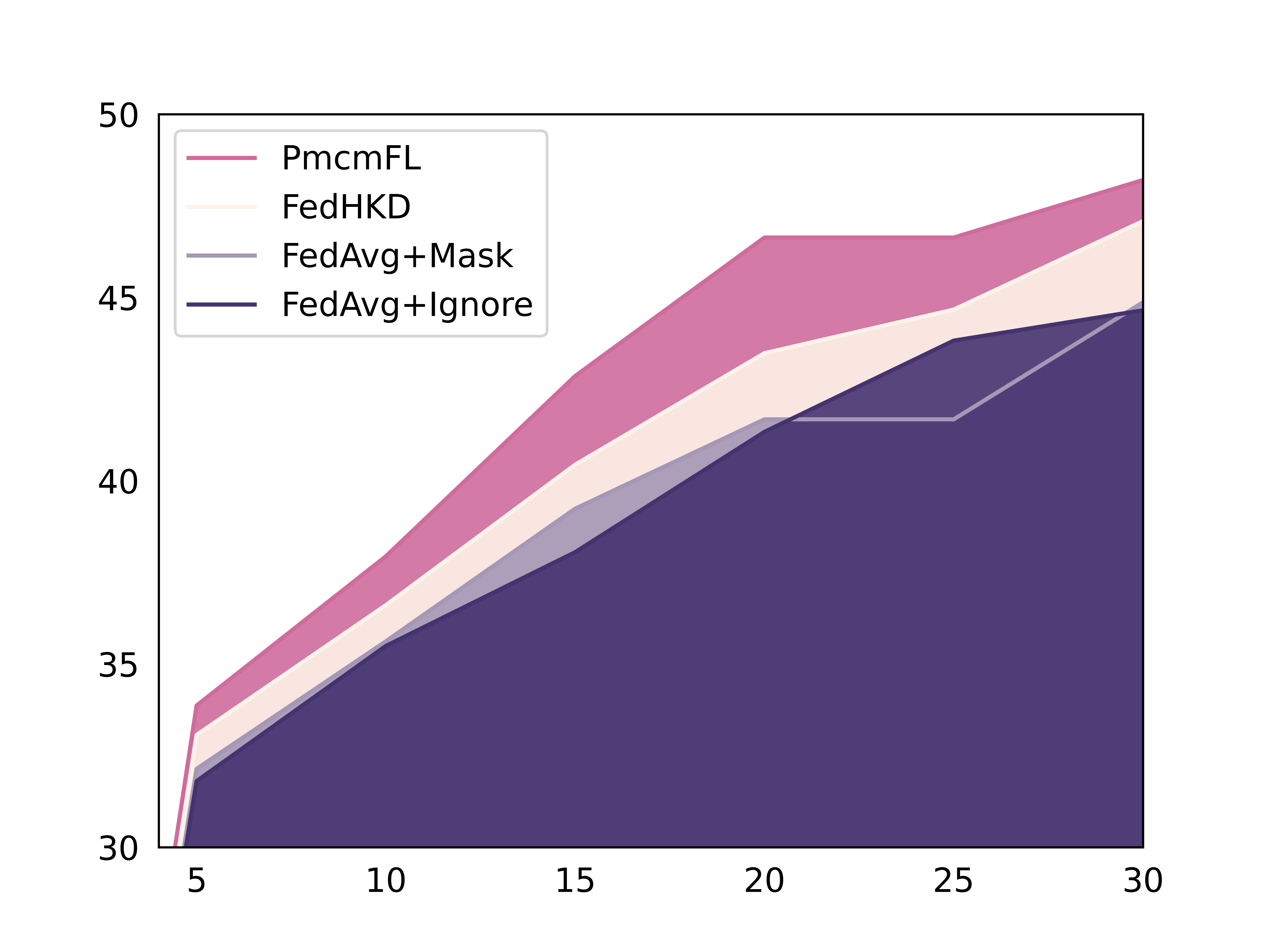}
    }
    \subfigure[Clients selection rates (\%)]{
        \includegraphics[width=0.46\linewidth]{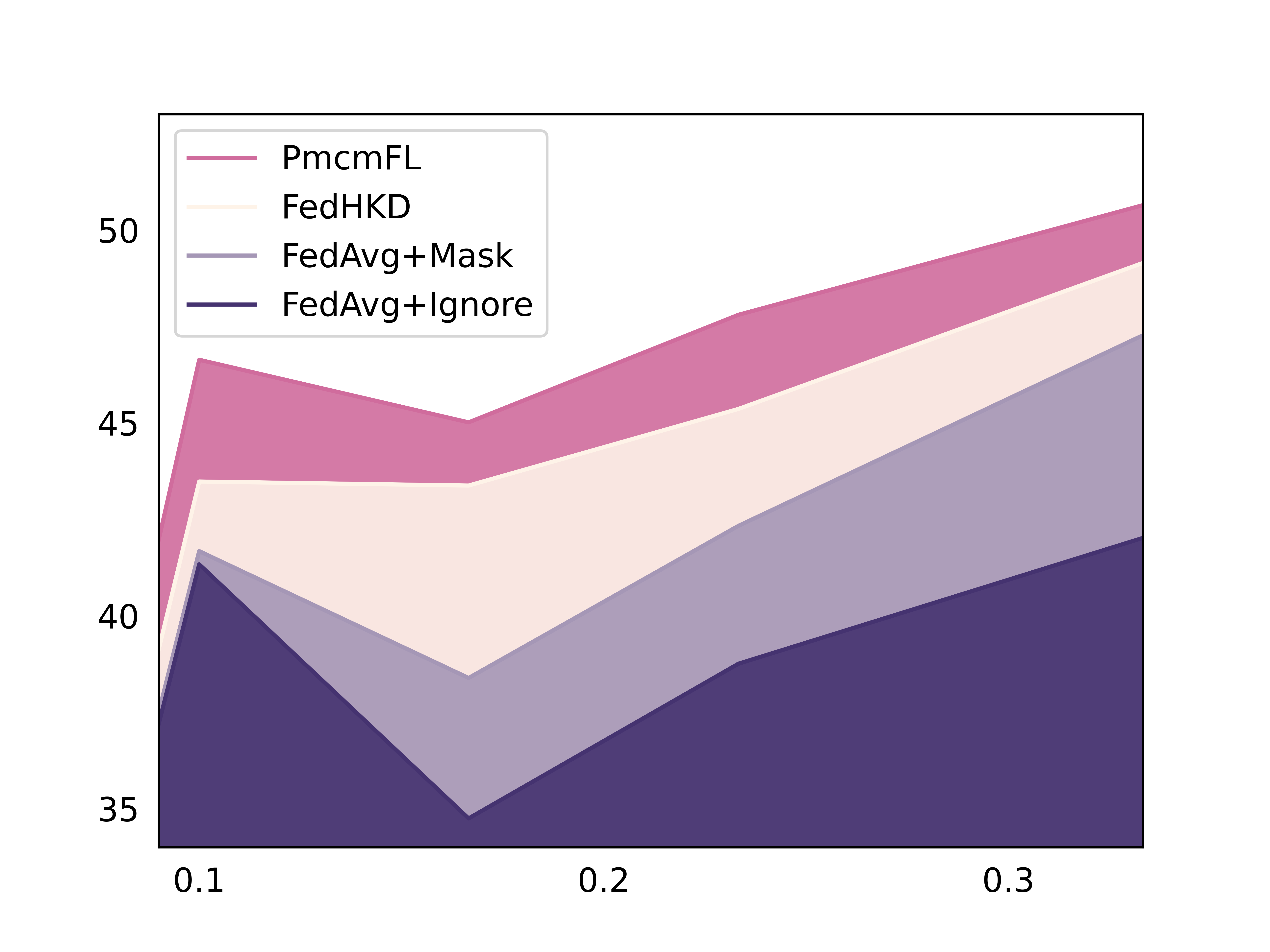}
    }
    \vspace{-3mm}
    \subfigure[Local training epochs]{
        \includegraphics[width=0.46\linewidth]{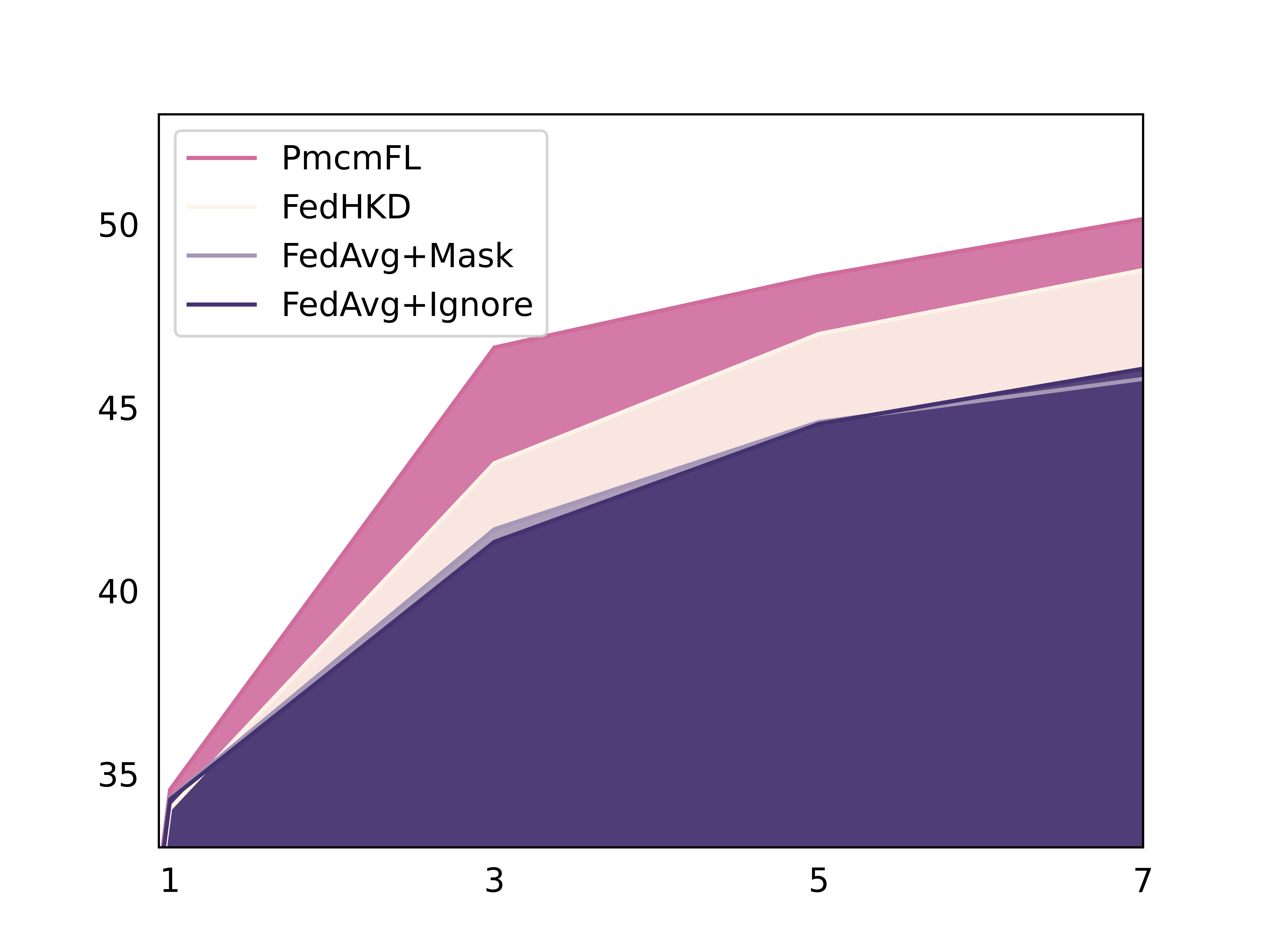}
    }
        \subfigure[Learning rates ($10^{-5}$)]{
        \includegraphics[width=0.46\linewidth]{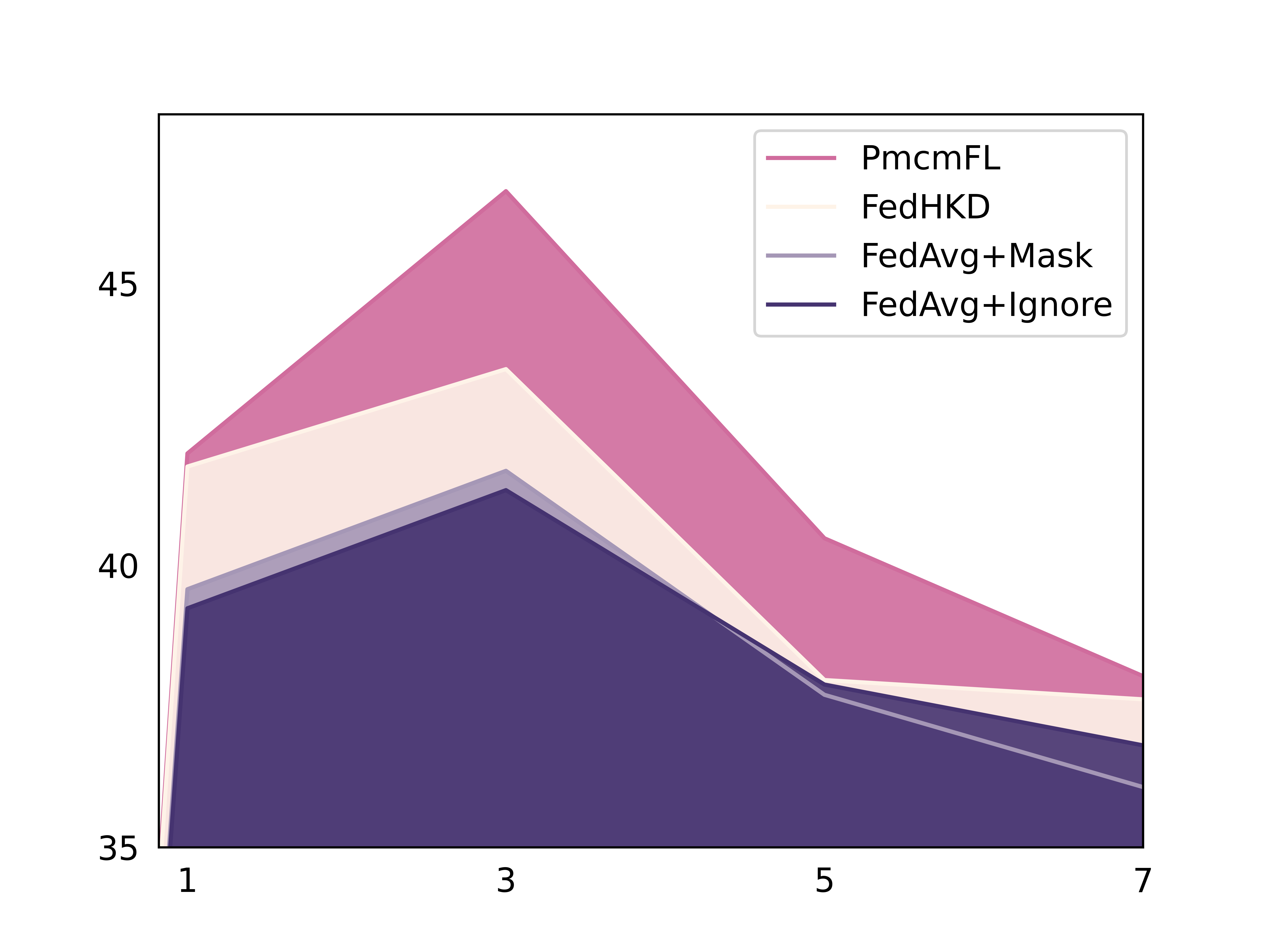}
    }    
    \end{center}
    \vspace{-3mm}
    \caption{Compare the performance of our PmcmFL with baseline frameworks under different FL settings.
    Our PmcmFL consistently achieves the best performance.}
    \vspace{-3mm}
\label{rubustness}
\end{figure}

Figure \ref{rubustness} illustrates the performance of various FL Frameworks across different communication rounds, client selection rates, local training epochs, and learning rates.
We conduct experiments with 50\% modality missing while keeping all federated settings constant except the one being investigated.
Four representative FL frameworks are compared: 1) FedAvg with ignoring modality-incomplete data; 2) FedAvg with Gaussian Random Mask; 3) FedHKD, the existing one with optimal performance; 4) our PmcmFL.
The experimental results demonstrate that, under different FL settings, our PmcmFL consistently achieved optimal performance, showcasing robustness to FL settings.

\section{Conclusion}
\label{sec:conclusion}

In this paper, we investigate multimodal federated learning with uncertain modality missing in both training and inference.
Our work introduces a prototype library to empower the FL framework with the capability to alleviate performance degradation resulting from modality missing.
Based on the prototype library, we construct a task-calibrated training loss, a model-agnostic unimodal inference strategy, and a proximal term.
The effectiveness of our framework has been validated through comparisons with state-of-the-art methods under various missingness settings.

% Our future work will focus on constructing a prototype library with richer semantics and extending Prototype Mask to a data augmentation method, or improving the matching accuracy from representations to prototypes to achieve better performance in inference with missing modalities.
% Moreover, using prototypes to handle missing or incorrect labels is also worth trying.

Our forthcoming efforts will center on riching semantics of prototypes, improving the prototype match accuracy, and extending Prototype Mask to a data augmentation method. 
Furthermore, we recognize the potential of using prototypes to tackle issues with missing or inaccurate labels, and we also intend to explore this avenue.

\section*{Impact Statements}
This paper presents work whose goal is to advance the field of Machine Learning, thereby facilitating the practical application of multimodal learning in real-world scenarios.
The dataset we use is publicly available and does not involve privacy issues.

% There are many potential societal consequences of our work, none which we feel must be specifically highlighted here.

\bibliographystyle{icml2024}
\bibliography{main}

\newpage
\appendix
\onecolumn

\section{Extend to Other Multimodal Tasks}
\label{appendix: extend}
In the main body, we illustrate the framework of PmcmFL using a text-image multimodal classification task (VQA) as an example and apply VQA to evaluate the framework's performance.
However, this does not imply that our PmcmFL is limited to specific modalities or tasks.

On the contrary, as the PmcmFL framework does not depend on specific model structures and only requires the model to be abstractly divided into modal encoders, fusion layers, and task heads, PmcmFL can be directly extended to almost all \textbf{multimodal classification tasks}.

In Table~\ref{other task}, we list some datasets to which PmcmFL can be extended, including the modalities and tasks associated with these datasets. The table represents just the tip of the iceberg for multimodal tasks that require modal fusion.

\begin{table}[h]
    \caption{Datasets that PmcmFL can be extended to.}
    \begin{center} 
        \begin{tabular}{ccccc}
             \toprule
             {Datasets} && {Modalities} && {Tasks} \\
             \midrule
              CrisisMMD        & &  Image, Text         & &  Crisis Information Classification \\
              Hateful-Memes    & &  Image, Text         & &  Hateful Content Detection \\
              MOSI             & &  Video, Audio, Text  & &  Multimodal Sentiment Analysis \\
              MELD             & &  Audio, Text         & &  Multimodal Sentiment Analysis \\
              UCF101           & &  Video, Audio        & &  Multimodal Action Recognition \\        
             \bottomrule
        \label{other task}
        \end{tabular}
    \end{center}
\end{table}

We can see that the datasets to which PmcmFL can be extended encompass inputs from various modalities and involve a variety of task types.

As for the extension strategy, taking the multimodal sentiment classification task on the MOSI dataset as an example, one simply needs to replace the image-text encoder described in PmcmFL with a video encoder, speech encoder, and text encoder. 
The fusion module can be specially designed for the task, along with a corresponding classification task head.

\section{Algorithm of PmcmFL}
\label{appendix: algorithm}

Algorithm~\ref{alg1} and Algorithm~\ref{alg2} illustrate the flowcharts of our PmcmFL during training and inference, respectively.

% PmcmFL training algorithm
\begin{algorithm*}[h]
    \caption{Training of PmcmFL.}
    \label{alg1}
    
    \KwIn{
        Number of federated communication rounds $T$,
        number of clients $C$,
        number of local epochs $E$,
        server model,
        local model $(\mathcal{E}_n,\mathcal{F}_n,\mathcal{G}_n)$,
        local learning rate $\eta$,
        dataset $\mathcal{D}_n=(\mathcal{M}_n,\mathcal{I}_n,\mathcal{T}_n)$ of the $n$-th client
        and fraction of clients $s$ that are selected to perform computation in each round.
    }
    
    \KwOut{The final server model's parameters $\omega^T$}
    
    \BlankLine

    \textbf{ServerExecutes}:

    \Indp
    
    Initialize ${\omega}^0$ randomly;
    
    Initialize global prototypes $\mathcal{P_I}_{-local}, \mathcal{P_T}_{-local}, \mathcal{P_F}_{-local}$ with zero tensors;
    
    \For{$t=1,2,\cdots,T $}
    {
        Compute global prototypes $\mathcal{P_I}, \mathcal{P_T}, \mathcal{P_F}$ according to Equation~\ref{equation04}~\ref{equation05}~\ref{equation06}; 

        $\mathcal{P_I}_{-local}, \mathcal{P_T}_{-local}, \mathcal{P_F}_{-local} \leftarrow [~], [~], [~]$;

        $S_t \leftarrow$ random set of $\max(s\cdot C, 1)$ clients;
        
        \For{\textnormal{each client $n$ in $S_t$}}{
            send the global model's parameters 
            $\omega^{t-1}$ and global prototypes $\mathcal{P_I}, \mathcal{P_T}, \mathcal{P_F}$ to client $n$; 

            $ \omega^t_n, \mathcal{P_I}_n, \mathcal{P_T}_n, \mathcal{P_F}_n \leftarrow $ 
            \textbf{ClientLocalTraining} 
            $(n,t,\omega^{t-1}, \mathcal{P_I}, \mathcal{P_T}, \mathcal{P_F})$; 

            $\mathcal{P_I}_{-local}, \mathcal{P_T}_{-local}, \mathcal{P_F}_{-local} \leftarrow \mathcal{P_I}_{-local} + \mathcal{P_I}_n, \mathcal{P_T}_{-local} + \mathcal{P_T}_n, \mathcal{P_F}_{-local} + \mathcal{P_F}_n$;
        } 
        Local models are aggregated into the global model:
        ${\omega}^t={\sum}_{n\in S_{t}} \frac{|\mathcal{D}_n|}{|\mathcal{D}|} {{\omega}^t_n}$
    }
    % \Indm
    % \BlankLine

    \textbf{ClientLocalTraining}
    $(n,t,{\omega}^{t-1}, \mathcal{P_I}, \mathcal{P_T}, \mathcal{P_F})$:

    \Indp
    \For{\textnormal{epoch $i=1,2,\cdots E$}}{
    Local update 
    $\omega^{(t,i)}_n \leftarrow \omega^{(t,i-1)}_n - \eta \nabla L(\mathcal{D}_n,\mathcal{P_I}, \mathcal{P_T}, \mathcal{P_F};\omega^{(t-1,i-1)}_n)$
    according to Equation~\ref{equation11}; 
    \;}

     Compute local prototypes $\mathcal{P_I}_n, \mathcal{P_T}_n, \mathcal{P_F}_n$ according to Equation~\ref{equation04}~\ref{equation05}~\ref{equation06};
    
    Return $\omega^t_n, \mathcal{P_I}_n, \mathcal{P_T}_n, \mathcal{P_F}_n$;
    
    \Indm
\end{algorithm*}

% PmcmFL inference algorithm
\begin{algorithm}[h]
    \caption{Inference of PmcmFL.}
    \label{alg2}
    
    \KwIn{
        Server model $(\mathcal{E},\mathcal{F},\mathcal{G})$,
        Testing set.
    }
    
    \KwOut{Labels for classification tasks}

    \BlankLine

    \If{\textnormal{input is an image-text pair $(i,t)$}}{
        \hspace{1mm} Compute $logits=\mathcal{G(F(E}(i,t)))$
    }
    
    \ElseIf{\textnormal{input is an image $i$}}{
        \hspace{1mm} Compute the image representation $h^{(i)}=\mathcal{E}(i)$; \\
        Search for the corresponding image prototype $\mathcal{P_I}^j$ using matching function $m(h^{(i)})$; \\
        Find the corresponding text prototype $\mathcal{P_T}^j$ through the association in the prototype library; \\
        Compute $logits=\mathcal{G(F(}h^{(i)},\mathcal{P_T}^j))$;
    }    
    \ElseIf{\textnormal{input is text $t$}}{
        \hspace{1mm} Compute the text representation $h^{(t)}=\mathcal{E}(t)$; \\
        Search for the corresponding text prototype $\mathcal{P_T}^j$ using matching function $m(h^{(t)})$; \\
        Find the corresponding text prototype $\mathcal{P_I}^j$ through the association in the prototype library; \\
        Compute $logits=\mathcal{G(F(}\mathcal{P_I}^j,h^{(t)}))$;
    }
    
    Calculate probability distribution and predict the label:
    $\hat{y}=\mathop{\textnormal{argmax}}\textnormal{Softmax}(logits)$
    
\end{algorithm}

\section{Details on Multiway Transformer}
\label{appendix: multiway transformer}
We will first introduce the differences between Multiway Transformers and regular Transformers here. We will also emphasize that although we use multi-head Transformers as encoders, any encoder architecture is applicable within the PmcmFL framework.

\subsection{Multiway Transformer}
In summary, a typical Transformer consists of Multi-Head Self-Attention layers (\textbf{Attention}), Feedforward Networks (\textbf{FFN}), Layer Normalizations, and Residual Connections.
The difference between Multiway Transformers and regular Transformers lies in the Attention layers and the FFNs. Specifically, the Attention layers in Multiway Transformer are modality-shared self-attention, and the FFNs are Modality-Specific Feedforward Networks (also known as Modality-Specific Experts Networks), meaning each modality has its own FFN.

We use images and text as examples to illustrate how Multiway Transformer switch encoding modes.
The input images and text are initially processed into sequences of tokens involving patchify or tokenization, together with positional encoding.
The initial token sequences (the image token sequence, the text token sequence, and the image-text token sequence) can be represented as:
\begin{equation}
    z^0_i=[z_{i0},z_{i1},\cdots,z_{iN_i}],
\end{equation}
\begin{equation}
    z^0_t=[z_{t0},z_{t1},\cdots,z_{tN_t}],
\end{equation}
\begin{equation}
    z^0=cat(z^0_i,z^0_t)=[z_{i0},\cdots,z_{iN_i},z_{t0},\cdots,z_{tN_t}],
\end{equation}
where $z_{i*}$ denotes the image token, $z_{t*}$ denotes the text token, $z_{i0}$ denotes the image CLS token, $z_{t0}$ denotes the text CLS token, $N_i$ and $N_t$ denote the number of image tokens and text tokens respectively.

For the Shared Self-Attention layer at $l$-th encoding layer, we represent it as ${Attention}^l(\cdot)$.
For the Modality-Specific Expert Networks at $l$-th encoding layer, we use ${FFN}^l_i(\cdot)$ and ${FFN}^l_t(\cdot)$ to respectively represent the vision expert network and the language expert network.
For the sake of clarity, we have omitted Residual Connections and Layer Normalizations.

When dealing with only image inputs, Multiway Transformer utilizes the attention module and the visual expert network at each encoding layer:
\begin{equation}
    z^l_i={{FFN}^l_i(Attention}^l(z^{l-1}_i)).
\end{equation}

When dealing with only text inputs, Multiway Transformer utilizes the attention module and the language expert network at each encoding layer:
\begin{equation}
    z^l_t={{FFN}^l_t(Attention}^l(z^{l-1}_t)).
\end{equation}

When dealing with image-text pairs, Multiway Transformer first employs the shared self-attention module for cross-modal interaction:
\begin{equation}
    a^l=cat(a^l_i,a^l_t)={Attention}^l(z^{l-1}).
\end{equation}
Subsequently, individual expert networks are applied to tokens from the corresponding modality:
\begin{equation}
    z^l=cat({FFN}^l_i(a^l_i),{FFN}^l_t(a^l_t)).
\end{equation}

By utilizing Modality-Specific Expert Networks, Multiway Transformers can switch to different encoders.

\subsection{Decoupling Framework and Model}

It is important to highlight that \textbf{PmcmFL is a universal framework and the encoders used in our PmcmFL are not limited to the Multiway Transformer}
In fact, our framework is flexible to incorporate with encoders of other architectures.

\begin{figure}[h]
    \begin{center}
    \includegraphics[width=0.4\linewidth]{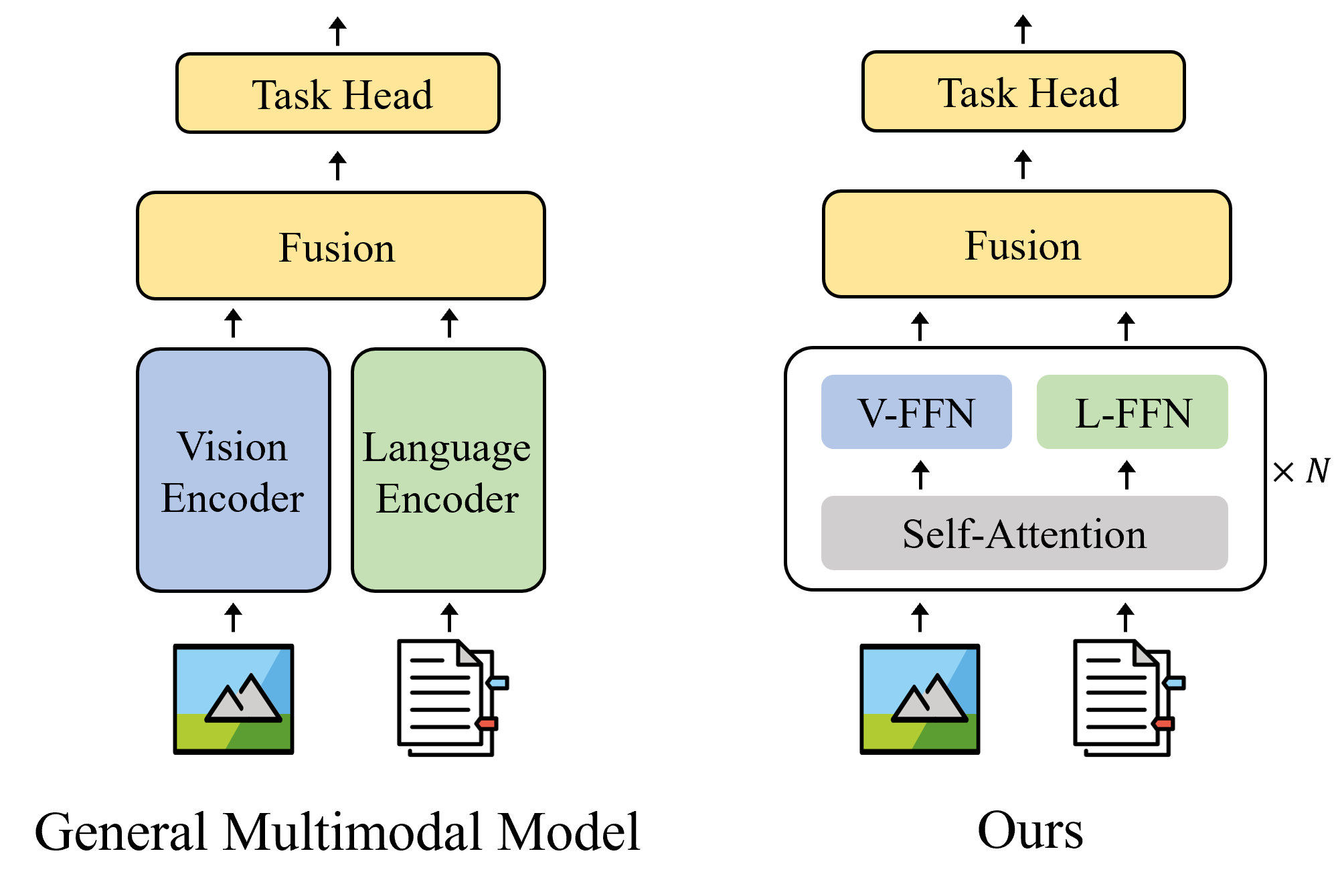}
    \caption{Comparison of general multimodal models and the multimodal model we used. 
    The main difference lies in the architecture of the encoder: we adopt the Multiway Transformer, whereas general multimodal models use modality-specific encoders.
    }
    \end{center}
    \label{model}
\end{figure}

Figure~\ref{model} illustrates the general architecture of multimodal models and the architecture we use. The same as ours, general multimodal models consist of three main parts: modality-specific encoders, a fusion module, and a task head. 

Due to the similar architecture, general multimodal models can be easily developed in our PmcmFL. To elaborate, we can construct a prototype library by utilizing the outputs of modality-specific encoders and the fusion module.
Based on the prototype library, PmcmFL can execute as usual.

It is worth noting that if the fusion module employs the Transformer architecture, it could be a wise move to integrate a module of bottleneck architecture after the modality-specific encoders to obtain low-dimension representation for computing prototypes, which will significantly reduce federated communication overhead. 

The reason we adopt Multiway Transformer is due to its superior ability in handling multimodal interaction. 
In our PmcmFL, Multiway Transformer not only functions as modality-specific encoders but also functions as dual-stream interactive encoders to achieve better modality fusion.

%%%%%%%%%%%%%%%%%%%%%%%%%%%%%%%%%%%%%%%% theory %%%%%%%%%%%%%%%%%%%%%%%%%%%%%%%%%%%%%%%%%%%
% \section{Generalization Error Bound}

%%%%%%%%%%%%%%%%%%%%%%%%%%%%%%%%%%%%%%%%%%%%%%%%%%%%%%%%%%%%%%%%%%%%%%%%%%%%%%%%%%%%%%%%%%%

\section{Details on the Dataset}
\label{appendix: dataset}
In this section, we provide a detailed description of the construction process for ting VQA, along with visualizations for non-IID data and modality missing.

\subsection{Tiny VQA}
Due to the typically slower convergence of models in federated learning, the training cost is significantly increased. 
The full scale of VQAv2 is too extensive for evaluating federated learning frameworks. 
Therefore, we evaluate on a \textbf{randomly selected subset} of VQAv2. 
This approach has been demonstrated in prior works, such as CreamFL.

Our tiny VQA is a scaled-down version, reducing both label quantity and training data volume to one-tenth of the original VQA task. 
The original VQA is commonly seen as a classification task with over 3,000 classes, supported by a vast training dataset exceeding 640,000 samples.
We refine the labels by excluding those with fewer than 180 occurrences, reshaping VQA into a classification challenge with 310 classes. 
For training, we randomly chose 64,000 image-text pairs, equivalent to one-tenth of the combined original VQA task training and validation sets. 
Additionally, our test set comprises 5,000 random image-text pairs randomly selected, ensuring none overlap with the training set. 
All experiments are conducted on ting VQA, except when compared with CreamFL.

\subsection{Visualization for Data Distribution}
To simulate real-world scenarios, we distributed the training data of tiny VQA across 30 clients, using a Dirichlet distribution with a hyperparameter of 0.1 as the basis for non-IID partitioning.
Previous works involving non-IID have almost adopted this hyperparameter setting, which describes a severe non-IID scenario.

Additionally, to simulate uncertain modality missing, we assigned a fixed missing rate for each modality on every client, thereby conforming to the Bernoulli distribution.
In our experiments, we considered five missing ratios: 10\%, 20\%, 30\%, 40\%, and 50\%. 

We visualized the data/modality distribution to illustrate the corresponding non-IIDness and modalities missingness as follows.

\begin{figure}[h]
    \begin{center}
    \includegraphics[width=0.75\linewidth]{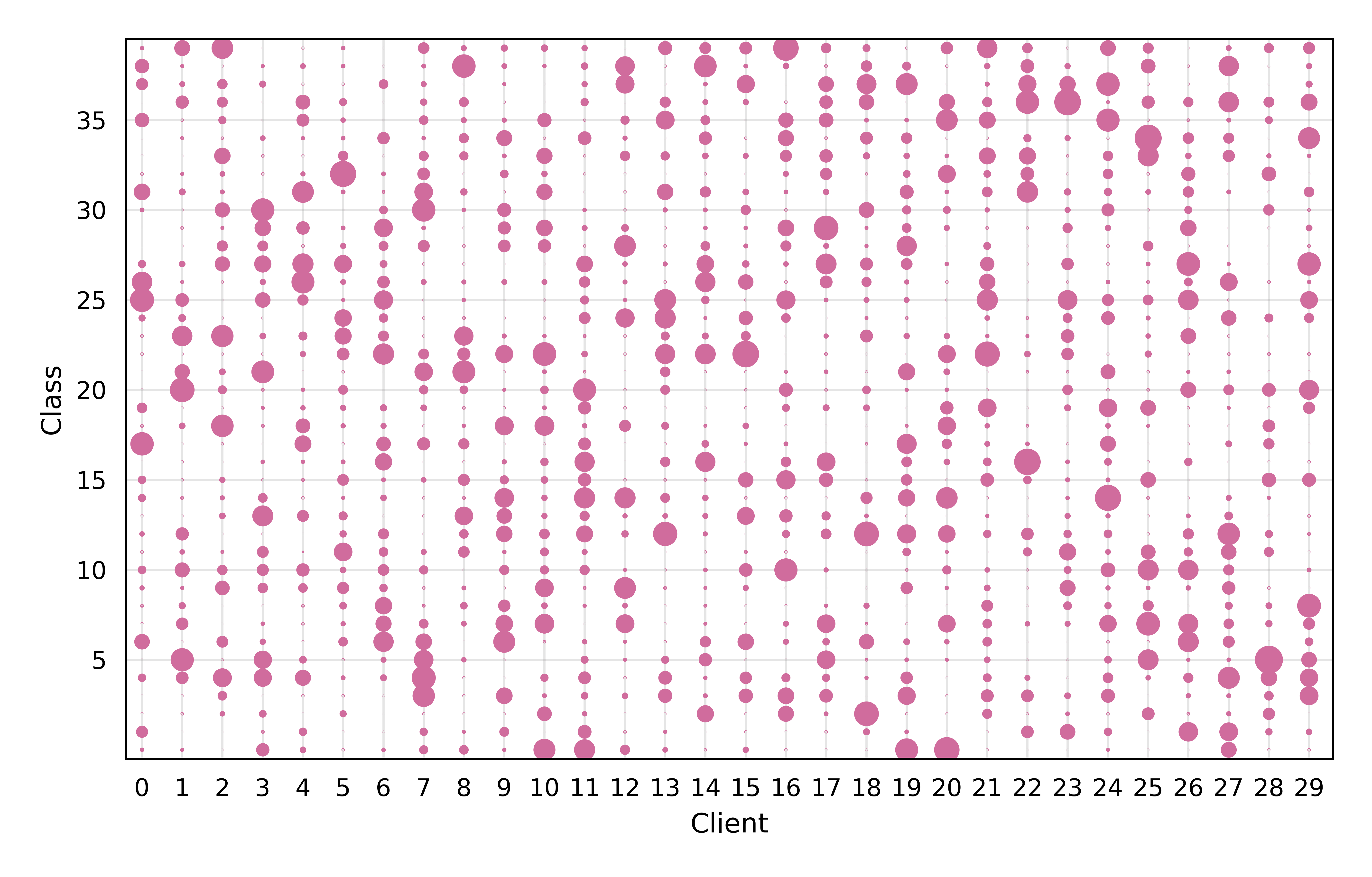}
    \caption{The data distribution of diverse classes on each client. The size of the dots represents the amount of data.}
    \end{center}
    \label{noniid}   
\end{figure}

\begin{figure}[h]
    \begin{center}
    \includegraphics[width=0.7\linewidth]{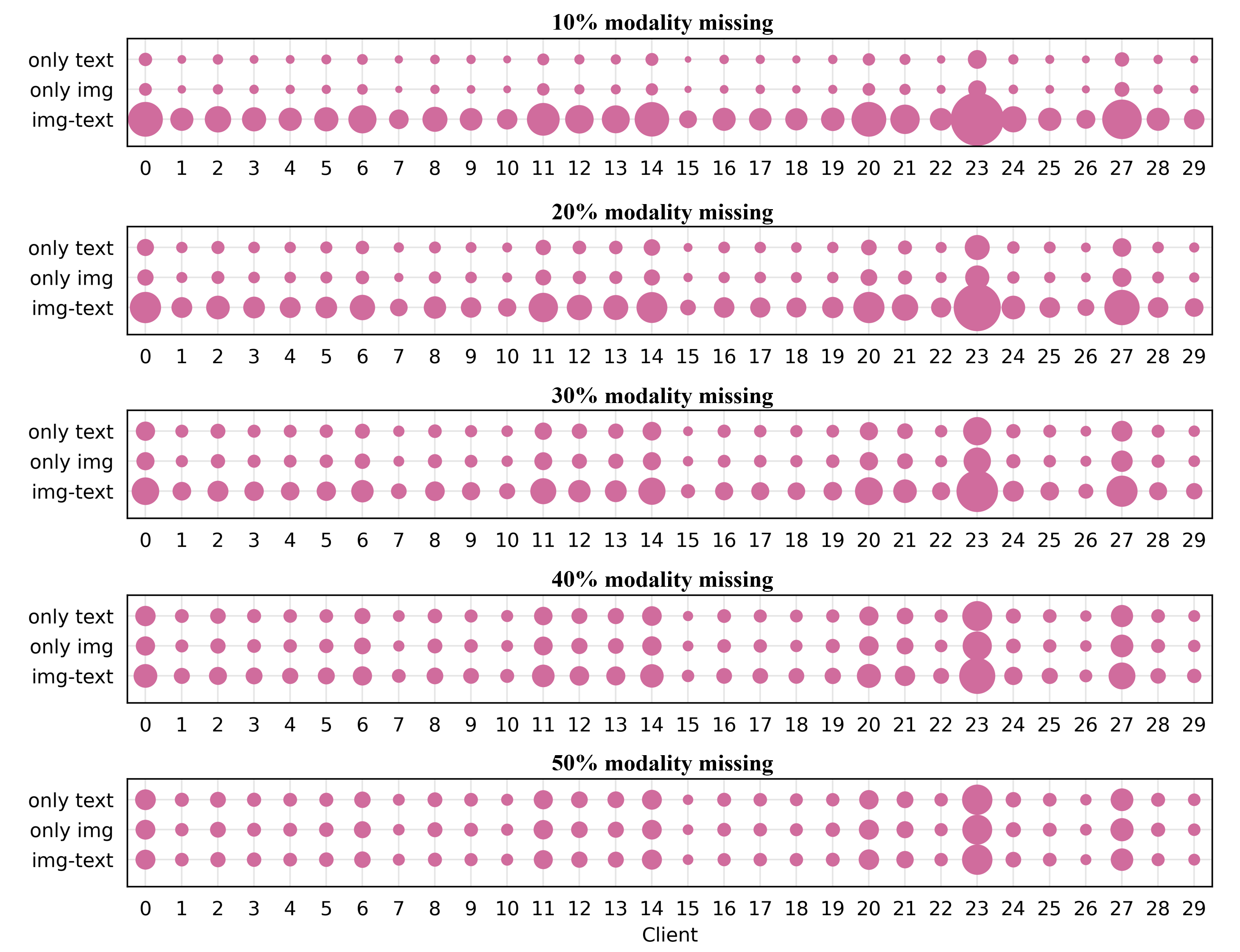}
    \caption{The modality distribution of various clients under different modality missing rates. 
    The size of the dots represents the amount of data.}
    \end{center}
    \label{missing}
\end{figure}

\textbf{Visualization for Non-IIDness.}
We selected 40 classes from a pool of 310, visualizing their distribution across various clients.
Figure~\ref{noniid} displays the diversified results of the visualization, indicating the heterogeneous class distribution across clients.

\textbf{Visualization for Modality Missingness.}
We visualized the modality distribution on various clients under different modality missing rates.
Figure~\ref{missing} displays the visualization results, showing the diversified modality distribution.

\section{Details on the Model}
\label{appendix: model}

In this section, we explain the model architecture used in our experiments.

For the Multiway Transformer encoder, we employ the architecture of BEIT3-base, which consists of 12 encoding layers. Detailed model architecture can be found in BEIT3's GitHub repository\footnote{BEIT3's GitHub repository: \url{https://github.com/microsoft/unilm/tree/master/beit3}}.
We make no modifications to the encoder architecture and utilize its pre-trained weight (pre-training data does not include VQAv2).
It is worth noting that when the input is an image-text pair, the original model only uses the image CLS token for subsequent tasks, whereas we utilize both the image CLS and the text CLS token. 
We simply add an additional output to accommodate our PmcmFL.

For the fusion module, we first apply layer normalization to the image CLS token and the text CLS token separately, then concatenate them. Subsequently, a linear layer and activation layer are applied to obtain the corresponding fusion representation.
The dimensions of both the CLS tokens and the fusion representations are 1×768.

For the task head, we use a linear layer to double the dimensions, followed by a layer normalization. 
After an activation layer, we then use another linear layer to reduce the dimensions to 310, which corresponds to the number of classification categories.

\section{Details on Baselines}
\label{appendix: baselines}

In this section, we elaborate on the details of the baselines.

\subsection{Principle of Selection}

For training with modality missing, we propose two innovations: Prototype Mask and Prototype Contrast. 
Prototype Mask is a strategy to address modality missing, while Prototype Contrast is a strategy to alleviate client heterogeneity and enhance the global model performance. 
Based on these, we aim to select the most suitable baseline.

\textbf{Handling modality missing.} \quad
When modality is missing, the simplest approach is to ignore the incomplete data and only use the data with complete modalities for training. We refer to this approach as \textbf{Ignore Missing}.
Previous works used zero tensors to replace the representation of the missing modality. We refer to this approach as \textbf{zero Mask}.
A straightforward extension is to use a random tensor following a Gaussian distribution for masking, which we refer to as \textbf{Random Mask}.
We consider these three strategies as baselines for handling the modality missing issue and compare them with our proposed Prototype Mask.

\textbf{Improving the global model.} \quad
Both non-independent and identically distributed (Non-IID) data and modality missing can lead to client heterogeneity, severely compromising the performance of the global model. 
\textbf{FedAvg}, as the fundamental federated learning algorithm, does not provide a solution to address non-IID issues. 
\textbf{FedProx} introduces a proximal term to constrain the updates of local models, demonstrating its effectiveness in mitigating the impact of non-IID scenarios. 
\textbf{FedIoT} tackles the problem from the perspective of model aggregation strategies, granting more aggregation weight to multimodal clients to alleviate the influence of modality-incomplete clients. 
Prototype-based Feature Alignment (\textbf{FA}) strategies in \textbf{FedPAC} and Prototype-based Hyper Knowledge Distillation (\textbf{HKD}) in \textbf{FedHKD} are both optimal approaches aimed at mitigating non-IID challenges.
We consider these five federated learning frameworks as baselines for improving the global model and compare them with our proposed Prototype Contrast.

\subsection{Implementation Details of Baselines}
We will elaborate on the implementation details of each baseline.

\textbf{FedAvg + Ignore.} \quad
Within the FedAvg framework, the training process involves disregarding all training data on clients with incomplete modalities and using only the data with complete modalities for training.

\textbf{FedMultimodal (FedAvg + Zero Mask).} \quad
Based on FedAvg, we adopt the Random Mask as the strategy for handling missing modalities.
This approach is consistent with FedMultimodal.

\textbf{FedMultimodal + RM (FedAvg + Random Mask).} \quad
Based on FedAvg, we adopt the Random Mask as the strategy for handling missing modalities.
Random tensors are obtained from a Gaussian distribution.

We compare three strategies for handling missing data under FedAvg and find that using the Random Mask yields the best results. Therefore, \textbf{subsequent baselines all adopt Random Mask}.

\textbf{FedProx + Random Mask.} \quad
Based on FedAvg, we adopt the Random Mask as the strategy for handling missing modalities.
According to the relevant paper, the weight of the proximal term is set to 1.0.

\textbf{FedIoT + Random Mask.} \quad
Based on FedAvg, we adopt the Random Mask as the strategy for handling missing modalities.
According to the paper of FedIoT, we set 100 times the aggregation weight for image-text pairs.

\textbf{FedPAC + Random Mask.} \quad
Based on FedAvg, we adopt the Random Mask as the strategy for handling missing modalities.
We implement a feature alignment strategy on the fused representations. 
According to the paper of FedPAC, the hyperparameter of feature alignment is set to 1.0.

\textbf{FedHKD + Random Mask.} \quad
Based on FedHKD, we adopt the Random Mask as the strategy for handling missing modalities.
Hyper knowledge distillation is applied in the fused representations
According to the paper of FedHKD, the hyperparameter of hyper knowledge distillation is set to 0.05.

\textbf{PmmFL + FA.} \quad
Based on FedAvg, we adopt the Prototype Mask as the strategy for handling missing modalities.
Feature alignment is applied to the fused representations, with the hyperparameter set to 1.0.

\textbf{PmmFL + HKD.} \quad
Based on FedAvg, we adopt the Prototype Mask as the strategy for handling missing modalities.
Hyper knowledge distillation is applied in the fused representations,  with the hyperparameter set to 0.05.

\textbf{CreamFL.} \quad
We use the results reported by CreamFL.

\section{Implementation Details}
\label{appendix: implementation}

We will go through the implementation details of each experiment one by one.

\textbf{Main Experiments.} \quad

The main experiment establishes consistent missing rates for each modality, specifically 0.1, 0.2, 0.3, 0.4, and 0.5.
We conducted testing using the modal complete test set and presented the results in Table~\ref{main results} (We conducted testing with modal partially missing data in Appendix~\ref{appendix:Inference with Partial Missing}.).
All experiment results are averaged under five runs with fixed random seeds.
We considered a conventional federated setting for the main experiment.
We set the communication rounds to 30. 
In each round, 5 clients are selected to participate in training (client selection rate is 0.167). 
We referred to the BEIT3 GitHub repository for the learning rate and optimizer parameters.
The learning rate is fixed at 3e-5. 
We use the AdamW optimizer with a weight decay set to 1e-4, a momentum of 0.9, and $\beta$ of [0.9, 0.98].
For the training loss, cross-entropy is used as the task loss, and the temperature for the prototype contrastive loss is set to 0.07, following the CLIP GitHub repository.
We only perform simple hyperparameter optimization for the weight of the prototype contrastive loss, selecting the optimal value from the set [5.0, 1.0, 0.5, 0.1, 0.01].
All experiments are conducted on 4 NVIDIA RTX 4090 GPUs using distributed data parallelism, with a total batch size of 48 (12 per GPU). 

\textbf{Ablation Studies.}
We conducted ablation studies on the Prototype Mask and Prototype Contrast in our PmcmFL.
The hyperparameter settings for the ablation experiments are the same as those for the main experiment. 
Note that Sec.~\ref{Robustness to federated settings} (Robustness to Federated Settings) can also be considered as an ablation study on federated settings.

\textbf{Inference with Missing Modalities.}
To intuitively demonstrate the role of Prototype Masks during the inference phase, we conduct inference with the complete missing of one modality.
The model used for this experiment is the best-performing model trained in the primary experiment with 30\% modality missing. 
Without loss of generality, we omit 100\% of the text modality during this experiment.
In this case, this experiment can also be regarded as an image unimodal inference experiment.

As described in the method section, we need to train a tiny model in the final round of federated communication to achieve the matching from representation to prototype.
For model-based matching, we trained tiny models on each client in the final round of federated communication.
The classification-based matching utilizes a 2.2M MLP architecture, accounting for less than 1\% of the communication round's overhead. 
The model takes image representations as input for a 310-class classification task and undergoes 100 epochs of training on each client, with a single GPU training time of under 1 minute. 
Similarly, the retrieval-based matching employs a 2.9M MLP architecture, constituting approximately 1\% of the communication round's overhead. 
This model takes image representations as input and generates corresponding image prototypes.
Its training involves an initial 200 epochs with contrastive loss, followed by 500 epochs of MSE loss.
The training time on a single GPU is approximately 3-5 minutes.

The server needs to select or integrate these tiny models from the clients.
We experimented with three strategies to explore the utilization of tiny models from various clients: selecting the model with the maximum client samples, model parameter aggregation through FedAvg, and model ensemble. 
The first strategy entails choosing the model from the client with the most data samples among the 30 clients. 
The model aggregation strategy combines the 30 small models through FedAvg. 
The model ensemble method merges classification probabilities or retrieval scores from individual models through weighted aggregation, with weights determined by the sample quantities of the respective clients.

\textbf{Qualitative Studies.}
We choose the optimal model trained in the main experiment with 50\% modality missing for quantitative experiments. 
Associated client models are also utilized.

To compare the heterogeneity among client models under FedAvg and PmcmFL, we randomly selected three models from five client models and visualized their fused representation distributions on the testing set.
To compare the performance of the global models under FedAvg and PmcmFL, we selected three categories (with corresponding labels 'red', 'blue', and 'green') from a total of 310 classes and visualized their fused representation distributions.

\textbf{Communication Efficiency.}
The experiments on communication efficiency utilize the results derived from the main experiment with 50\% modality missing.

\textbf{Robustness for federated setting.}
This experiment can also be considered as an ablation study on federated settings.
We test PmcmFL's robustness for varying communication rounds, client selection rates, local training epochs, and learning rates. 
To evaluate the robustness for communication rounds, we set communication rounds to [5, 10, 15, 20, 25, 30], with a client selection rate fixed at 0.1, local training epochs set to 3, and a fixed learning rate of 3e-5.
To evaluate the robustness for client selection rates, we set client selection rates to [0.1, 0.167, 0.233, 0.333], with communication rounds of 20, local training epochs set to 3, and a fixed learning rate of 3e-5.
To evaluate the robustness for local training epochs, we set local training epochs to [1, 3, 5, 7], with communication rounds of 20, a client selection rate fixed at 0.1, and a fixed learning rate of 3e-5.
To evaluate the robustness for learning rates, we set learning rates to [1e-5, 3e-5, 5e-5, 7e-5], with communication rounds of 20, a client selection rate fixed at 0.1, and local training epochs set to 3.

\section{Inference with Partial Modality Missing}
\label{appendix:Inference with Partial Missing}

We train the model with modalities missing at rates of [10\%, 20\%, 30\%, 40\%, 50\%], and performed inference on test sets with modalities missing at the same rates. 

In baselines, Random Mask serves as the strategy for handling missing modalities, except for FedMultimodal (FedAvg + Zero Mask), which employs Zero Mask.
In our PmcmFL, the Prototype Mask is used for handling missing modalities.

The experimental results are shown in Table~\ref{table: Inference with Partial Missing}. 
We compare our PmcmFL with existing approaches, so the two variants of our framework (PmmFL+FA and PmmFL+HKD) are not presented in the table.

As can be seen, the experimental results demonstrate the superiority of our PmcmFL.

\begin{table}
  \caption{Comparison of PmcmFL with baselines under various missing rates. 
  Modality missing occurs in both the training and testing phases. In our approach, the Prototype Mask is also used during the inference phase. $\rho_{train}$ and $\rho_{test}$ denote the missing rates in the training phase and testing phase, respectively. We bold the best results under the same training and test missing rate.}
\begin{center}
    \begin{tabular}{@{}l|c|cccccc@{}}
    \toprule
    \multirow{2}{*}{Methods} & \multirow{2}{*}{$\rho_{test}$} & \multicolumn{5}{c}{$\rho_{train}$} & \multirow{2}{*}{Sum}\\
    \cline{3-7}
    & & {10\%} & {20\%} & {30\%} & {40\%} & {50\%} & \\
    \midrule

    \multirow{5}{*}{FedAvg + Ignore}
    & 10\%               & 52.016 & 50.756 & 46.812 & 42.254 & 36.720 & 228.558  \\
    & 20\%               & 47.904 & 47.318 & 43.198 & 39.386 & 34.278 & 212.084  \\
    & 30\%               & 42.490 & 43.138 & 39.250 & 36.340 & 31.802 & 193.020  \\
    & 40\%               & 38.938 & 40.462 & 36.964 & 34.218 & 30.040 & 180.622  \\ 
    & 50\%               & 33.760 & 36.180 & 33.406 & 30.492 & 26.926 & 160.764  \\
    \midrule

    \multirow{5}{*}{FedMultimodal}
    & 10\%                 & 52.822 & 50.288 & 47.234 & 44.478 & 39.766 & 234.588  \\
    & 20\%                 & 50.310 & 47.624 & 45.056 & 42.280 & 38.366 & 223.636  \\
    & 30\%                 & 47.252 & 44.832 & 43.174 & 40.288 & 37.242 & 212.788  \\
    & 40\%                 & 45.496 & 43.034 & 41.904 & 38.824 & 36.228 & 205.486  \\
    & 50\%                 & 42.588 & 39.828 & 40.034 & 36.872 & 34.658 & 193.980  \\
   \midrule

    \multirow{5}{*}{FedMultimodal + RM}
    & 10\%                  & 52.268 & 50.242 & 46.230 & 45.352 & 39.962 & 234.054  \\
    & 20\%                  & 49.176 & 47.602 & 43.954 & 43.368 & 38.460 & 222.560  \\
    & 30\%                  & 45.290 & 44.862 & 41.614 & 41.120 & 36.966 & 209.852  \\
    & 40\%                  & 42.554 & 42.958 & 39.844 & 39.862 & 35.870 & 201.088  \\
    & 50\%                  & 38.568 & 39.964 & 37.348 & 37.536 & 34.082 & 187.498  \\
    \midrule

    \multirow{5}{*}{FedProx + Mask}
    & 10\%                & 50.382 & 47.392 & 44.790 & 45.234 & 39.940 & 227.738  \\
    & 20\%                & 48.094 & 44.978 & 42.848 & 43.170 & 38.462 & 217.552  \\
    & 30\%                & 45.690 & 42.592 & 40.808 & 41.514 & 36.936 & 207.540  \\
    & 40\%                & 44.300 & 41.002 & 39.414 & 40.240 & 35.806 & 200.762  \\
    & 50\%                & 42.132 & 38.350 & 37.372 & 38.136 & 34.138 & 190.128  \\
    \midrule

    \multirow{5}{*}{FedIoT + Mask}
    & 10\%                 & 52.714 & 49.910 & 45.698 & 45.664 & 40.604 & 234.590  \\
    & 20\%                 & 50.278 & 47.112 & 43.544 & 43.486 & 39.130 & 223.550  \\
    & 30\%                 & 47.572 & 44.694 & 41.266 & 41.456 & 37.430 & 212.418  \\
    & 40\%                 & 45.606 & 43.016 & 39.726 & 39.968 & 36.226 & 204.542 \\
    & 50\%                 & 43.164 & 40.000 & 37.420 & 37.494 & 34.430 & 192.508  \\
    \midrule

    \multirow{5}{*}{FedPAC + Mask}
    & 10\%                 & 52.044 & 48.640 & 45.266 & 43.950 & 39.142 & 229.042  \\
    & 20\%                 & 49.738 & 46.094 & 43.266 & 41.934 & 37.618 & 218.650  \\
    & 30\%                 & 47.082 & 43.558 & 41.084 & 39.692 & 36.282 & 207.698  \\
    & 40\%                 & 45.402 & 42.024 & 39.656 & 38.258 & 35.376 & 200.716  \\
    & 50\%                 & 42.430 & 39.236 & 37.460 & 35.840 & 33.872 & 188.838  \\
    \midrule

    \multirow{5}{*}{FedHKD + Mask}
    & 10\%                 & 53.774 & 51.708 & 48.298 & 46.060 & 42.256 & 242.096 \\
    & 20\%                 & 51.222 & 49.654 & 45.826 & 44.004 & 40.946 & 231.652 \\
    & 30\%                 & \textbf{48.232} & \textbf{47.498} & 43.186 & 42.064 & 39.654 & 220.634  \\
    & 40\%                 & \textbf{46.524} & 45.306 & 41.590 & 40.962 & 37.908 & 212.290  \\
    & 50\%                 & \textbf{43.632} & \textbf{43.358} & 38.790 & 38.522 & 37.394 & 201.696  \\
    \midrule

    \multirow{5}{*}{PmcmFL(\textbf{Ours})}
    
    & 10\%     & \textbf{53.810} & \textbf{52.846} & \textbf{49.710} & \textbf{47.206}& \textbf{45.106} & \textbf{248.678} \\
    & 20\%     & \textbf{51.552} & \textbf{50.606} & \textbf{47.584} & \textbf{45.428}& \textbf{43.300} & \textbf{238.470} \\
    & 30\%     & 48.148 & 47.452 & \textbf{44.916} & \textbf{42.932}& \textbf{41.054} & \textbf{224.502} \\
    & 40\%     & 45.792 & \textbf{45.812} & \textbf{42.874} & \textbf{41.014}& \textbf{39.404} & \textbf{214.896} \\
    & 50\%     & 43.490 & 43.018 & \textbf{39.912} & \textbf{39.132}& \textbf{38.264} & \textbf{203.816} \\
    
    \bottomrule
    \end{tabular}
  
\end{center}
\label{table: Inference with Partial Missing}
\end{table}

%%%%%%%%%%%%%%%%%%%%%%%%%%%%%%%%%%%%%%%%%%%%%%%%%%%%%%%%%%%%%%%%%%%%%%%%%%%%%%%%%

\section{Compare with CreamFL}
\label{appendix: creamfl}
We compare PmcmFL with CreamFL, and due to the different task scenarios from the main experiment, we present the results here.

Similar to CreamFL, we conducted experiments in simple scenarios where there are only unimodal clients and modality-complete multimodal clients.
We train models on over 3000 of the most frequent answers (the original VQA task) and report the inference accuracy on 5K testing samples.
Table~\ref{compare with creamfl} shows the results.

\begin{table}[h]
  \begin{center}
\caption{Comparison of PmcmFL with baselines in specific task scenarios.}
  \begin{tabular}{@{}lc@{}}
    \toprule
    Methods & Accuracy \\
    \midrule
    FedAvg & 52.54 \\
    FedIoT & 53.06 \\
    FedMD & 57.43 \\
    FedET & 59.90 \\
    FedGEMS & 60.23 \\
    reamFL+Avg & 58.64 \\
    reamFL+IoT & 59.64 \\
    CreamFL & 62.12 \\
    PmcmFL(\textbf{Ours}) & \textbf{65.53} \\
    \bottomrule
  \end{tabular}
    \end{center}
  \label{compare with creamfl}
\end{table}

The results show our framework also achieves SOTA performance (a 3.38\% boost compared with CreamFL). 
This suggests that PmcmFL is equally applicable to specific task scenarios considered by previous works and outperforms methods specifically designed for such scenarios.

We believe that this performance improvement mainly stems from the utilization of the Multiway Transformer encoder, which excels in modality fusion and thus benefits the task.
This reflects the advantage of our federated framework compared with CreamFL, which cannot train the modality fusion module on the client.

\section{Discussions on ProtoMix}
\label{appendix: protomix}

We first present the testing set matching accuracy for each matching method, followed by an explanation of the results generated by ProtoMix.

\subsection{Accuracy of Various Matching Strategies}

Table~\ref{matching_acc} displays the matching accuracy of each matching strategy.
The data presented corresponds directly to Table~\ref{unimodal inference}.

\begin{table}[h]
  \begin{center}
  \caption{The matching accuracy of various matching strategies under top-1, top-5, top-10, and top-20 matches.}
  \begin{tabular}{@{}lcccc@{}}
    \toprule
    \multirow{2}{*}{Methods} & \multicolumn{4}{c}{Matching Accuracy (\%)} \\
    \cline{2-5}
    & {Top1} & {Top5} & {Top10} & {Top20} \\
    \midrule
    L1 Metric       &  2.46 &  5.94 & 10.78 & 28.46 \\ 
    L2 Metric       &  1.04 &  3.14 &  4.94 & 11.86 \\
    COS Metric      &  1.04 &  3.32 &  6.84 & 27.04 \\
    Classifier(max) & 19.10 & 46.28 & 50.80 & 55.16 \\
    Classifier(ens) & 22.34 & 54.38 & 62.72 & 69.14 \\
    Classifier(avg) & 23.20 & 51.16 & 57.18 & 62.74 \\
    MLP-Piror(max)   & 19.56 & 41.92 & 42.62 & 43.46 \\
    MLP-Piror(ens)   & 21.70 & 47.92 & 52.38 & 56.16 \\
    MLP-Piror(avg)   &  0.36 &  4.32 &  5.36 &  9.72 \\
    \bottomrule
  \end{tabular}
  \end{center}
  \label{matching_acc}
\end{table}

We can see that there is a significant difference in the matching accuracy of prototype matching strategies with different approaches. 
The top-1 matching accuracy of the model-free matching strategy is only around 1.04-2.46\%, while the top-1 prototype matching accuracy of the model-based matching strategy is approximately 19.10-23.20\% (except for MLP-Prior with parameter aggregation).

In addition, compared to the top-1 matching accuracy, the matching accuracy for top-k (k=5, 10, 20) shows significant improvement. 
This suggests that considering more prototypes may lead to obtaining more closely matched prototypes, which forms the basis for our proposal of Protomix.

Compared to Table~\ref{unimodal inference} in the main text, we observe a positive correlation between top-1 prototype matching accuracy and accuracy of inference with missing modalities.
The positive correlation is reasonable and the effect we desire, which also indicates that, for higher accuracy of inference with missing modalities, we should strive to improve the accuracy of prototype matching.
In fact, when we perform unimodal inference using the corresponding prototypes (prototype matching accuracy is 100\% ), its accuracy can reach 43.314\%, which is only 8.942\% lower than the modality-complete inference.

\subsection{Analysis of ProtoMix}
\begin{figure}[!t]
    \begin{center}
    \includegraphics[width=0.75\linewidth]{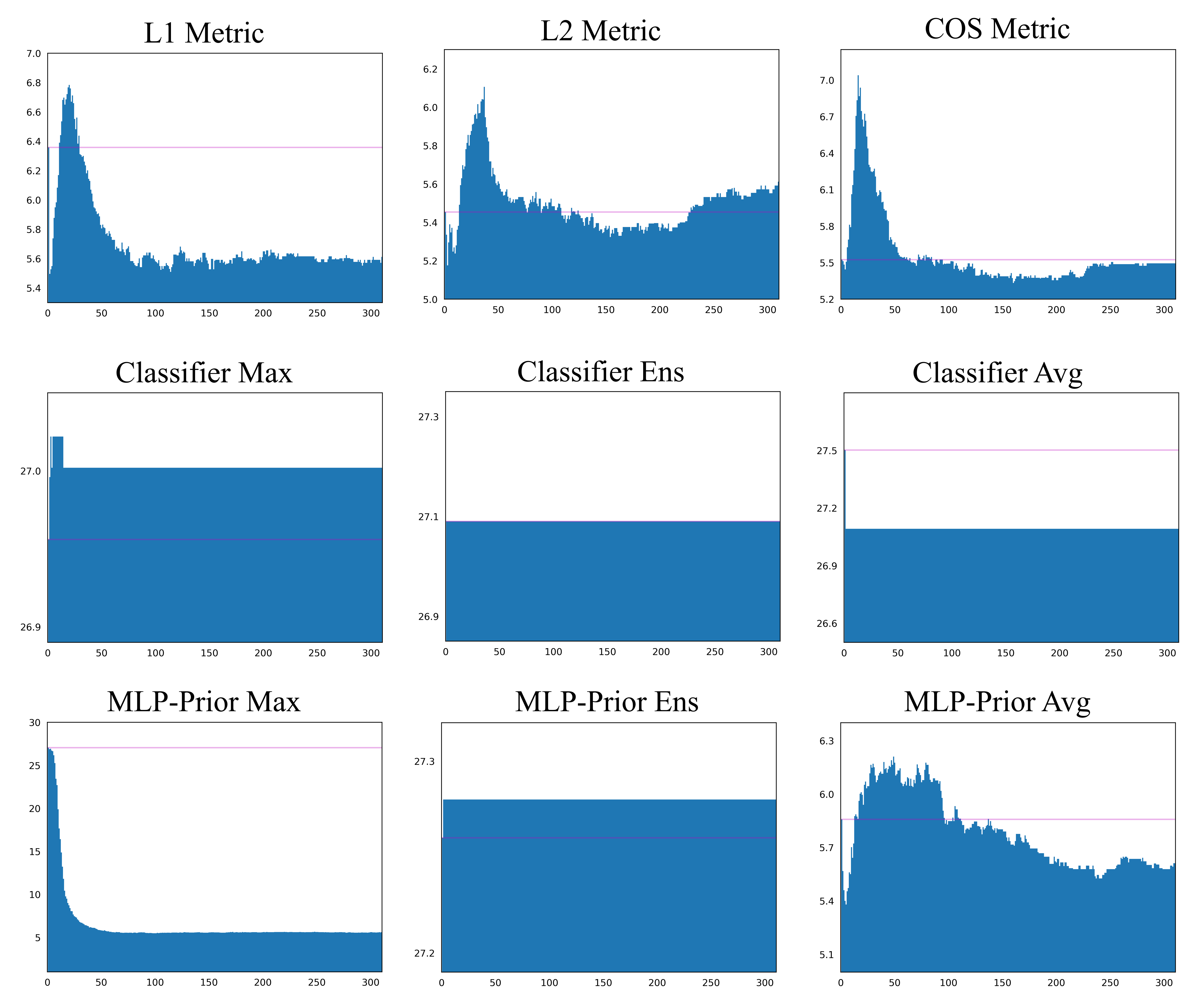}
    \caption{All possible ProtoMix results for nine matching strategies. 
    The horizontal axis represents the k values in the top-k ProtoMix, while the vertical axis represents the corresponding accuracy of inference with missing modalities.
    The horizontal line represents the accuracy of the top-1 ProtoMix.}
    \end{center}
    \label{fig:protomix}
\end{figure}

Figure~\ref{fig:protomix} displays all possible ProtoMix results for nine matching strategies.

We observe that almost all matching strategies benefit from ProtoMix, indicating that our proposed strategy of augmenting prototype semantics is effective. 
However, the performance improvement brought by Top-k ProtoMix is very limited and does not correspond to the accuracy of Top-k prototype matching, which suggests that there is still significant space for improvement in our ProtoMix.

Additionally, we observed significant differences in the performance of ProtoMix under different matching strategies, and we will analyze them one by one.
There are two key factors influencing the performance of ProtoMix: prototype matching accuracy and matching confidence.

For the three model-free matching strategies, both matching accuracy and confidence are relatively low. 
The results indicate that combining top-k prototypes with confidence-weighted fusion results in prototypes with more accurate semantics.

For the three classifier-based matching strategies, their matching accuracy is generally moderate, but the confidence in the top-1 match is exceptionally high. 
We observed that both a single classifier and a parameter-aggregated classifier have 19.10\% matching accuracy but almost boast over 99\% confidence in the top-1 class. 
The results in very little semantic information from other prototypes were obtained when combining top-k prototypes with confidence-weighted fusion. 
Consequently, the inference accuracy remains the same regardless of how many prototypes are fused. 
This limitation stems from the poor generalization ability of the classifier, as it tends to overfit the uneven distribution of client data during training on the client side.

For the three retrieval-based matching strategies, we observed that a single prior has 19.56\% prototype matching accuracy and possesses a top-1 retrieval score with less prominent confidence, creating a difference between the maximum prior and the ensemble prior. 
However, since parameter aggregation does not apply to non-classification tasks, the aggregated model exhibits low matching accuracy and confidence, leading to performance comparable to model-free matching strategies.

\section{Additional Visualization}

\subsection{Representation Distribution among Clients}
\begin{figure}[h]
    \begin{center}
    \includegraphics[width=0.95\textwidth]{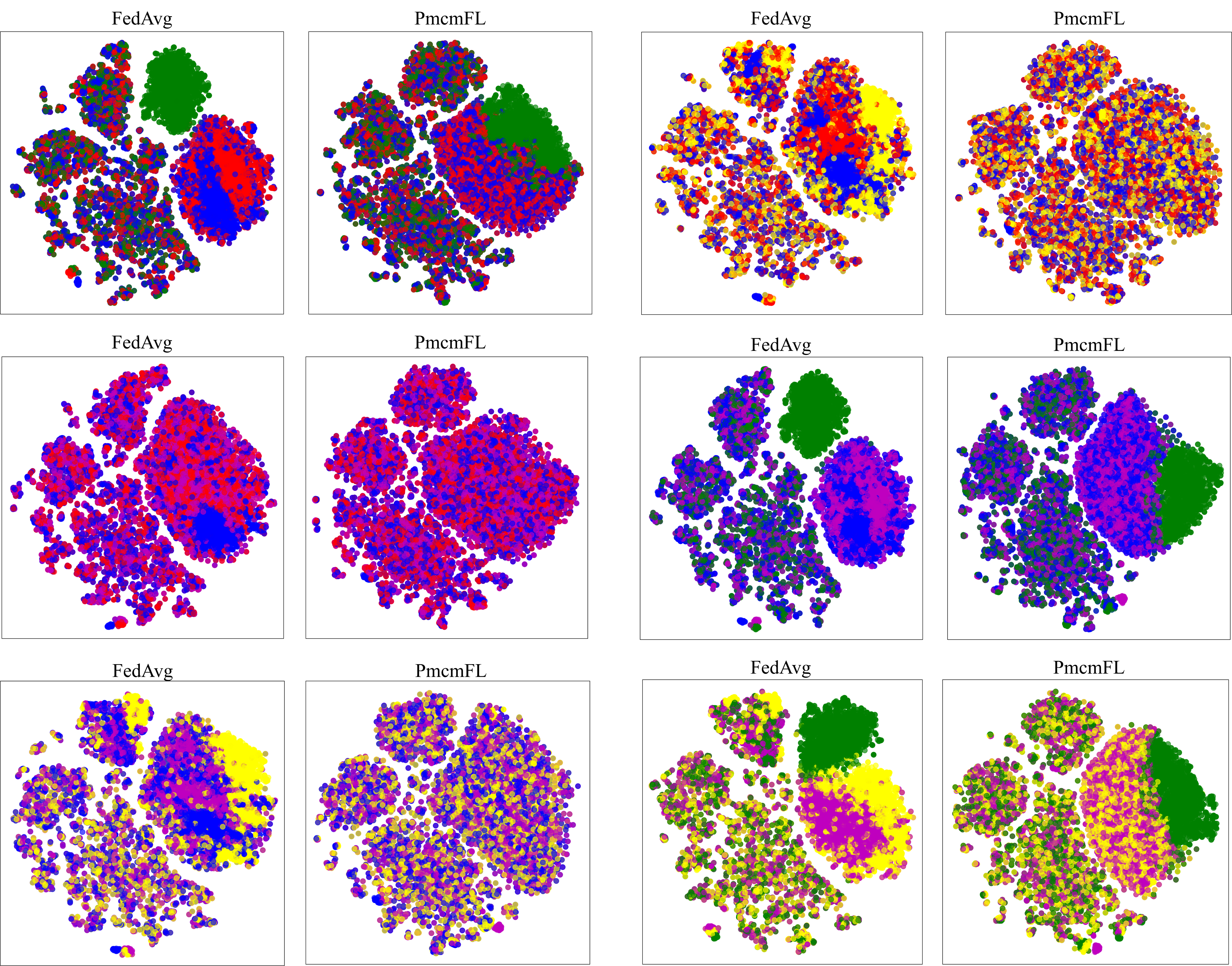}
    \caption{Representation Distribution among Clients.
    Dots of the same color come from the same client. Compared to the baseline, our PmcmFL reduces heterogeneity between clients.}
    \end{center}
    \label{tsne1-suppl}
\end{figure}

We visualized more representation distributions learned by various clients.
Figure~\ref{tsne1-suppl} displays the visualized results.
We observe that, compared to the baseline method FedAvg, our PmcmFL reduces the heterogeneity in the feature space among clients (i.e., mitigates client drift).

\subsection{Representation Distribution among Classes}
\begin{figure}[h]
    \begin{center}
    \includegraphics[width=0.95\textwidth]{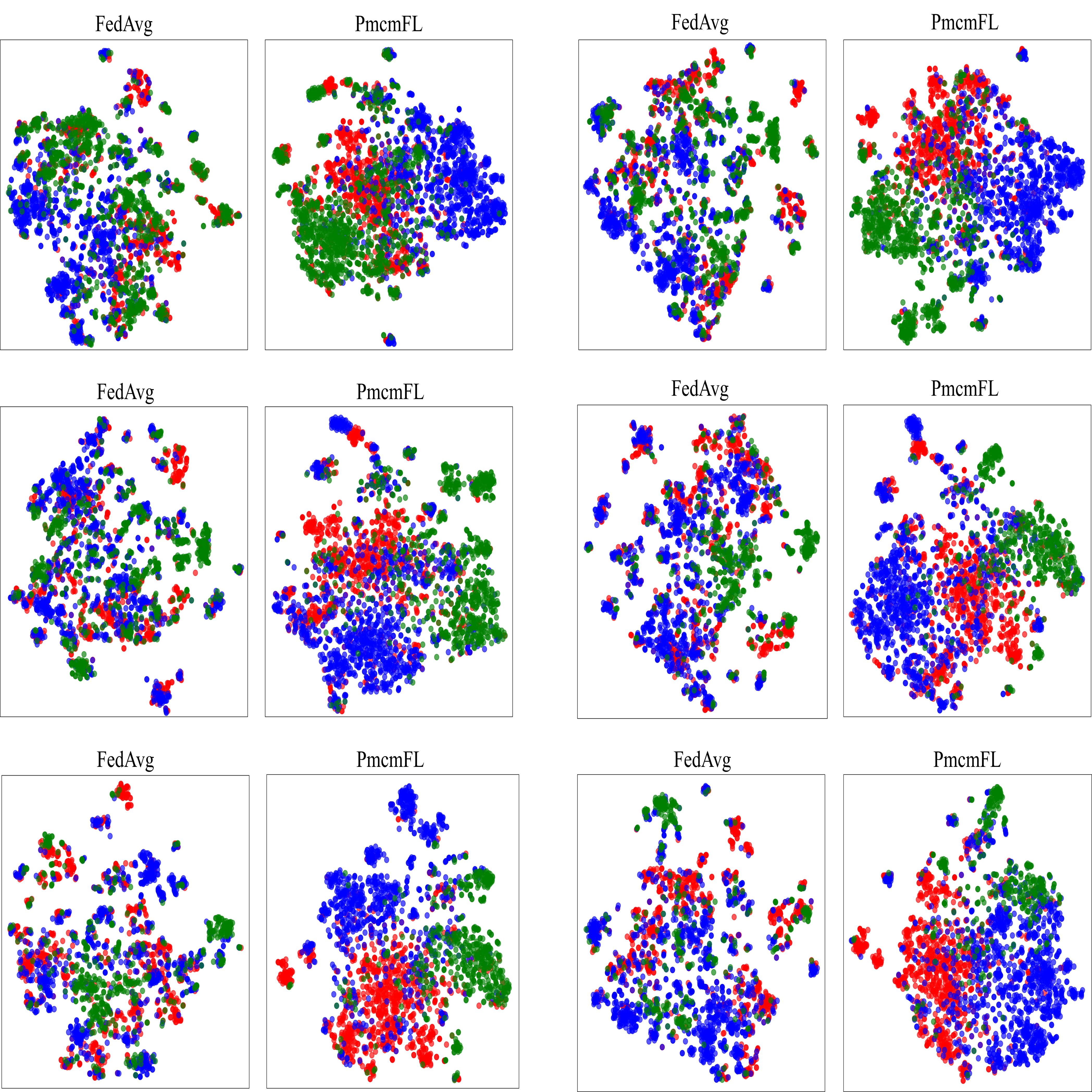}
    \caption{Representation Distribution among Classes.
    Dots of the same color represent the fused representation of the same class. Compared to the baseline, our PmcmFL can cluster representations by class.}
    \end{center}
    \label{tsne2-suppl}
\end{figure}

We visualized more representation distributions learned by the global model among various classes.
Figure~\ref{tsne2-suppl} displays the visualized results.
We can observe that, compared to the baseline method, our PmcmFL can cluster representations by class.

% \section{Future Work}

\end{document}